\documentclass[10pt,journal,compsoc]{IEEEtran}
\usepackage{multirow}

\ifCLASSOPTIONcompsoc
  \usepackage[nocompress]{cite}
\else
  \usepackage{cite}
\fi
\ifCLASSINFOpdf
  \usepackage[pdftex]{graphicx}
  \graphicspath{{./images/}}
  \DeclareGraphicsExtensions{.pdf,.jpeg,.png}
\else
\fi
\usepackage{amsmath}
\usepackage{algorithmic}

\ifCLASSOPTIONcompsoc
 \usepackage[caption=false,font=footnotesize,labelfont=sf,textfont=sf]{subfig}
\else
 \usepackage[caption=false,font=footnotesize]{subfig}
\fi
\newif\ifsubmit
\submitfalse

\ifsubmit
    \newcommand{\utkarsh}[1]{}
    
\else
    \usepackage{xcolor}
    \newcommand{\utkarsh}[1]{[{\color{cyan}US: #1}]}

\fi
\usepackage{soul}
\renewcommand\hl[1]{#1}

\begin{document}
%
\title{Hardware/Software co-design with ADC-Less In-memory Computing Hardware for Spiking Neural Networks}
%
%
%
%

\author{
        Marco~P.~E.~Apolinario,~\IEEEmembership{Student~Member,~IEEE,}~Adarsh~Kumar~Kosta,
        ~Utkarsh~Saxena,~\IEEEmembership{Student Member,~IEEE,}
        ~and~Kaushik~Roy,~\IEEEmembership{Fellow,~IEEE}
\IEEEcompsocitemizethanks{\IEEEcompsocthanksitem All authors are with the Elmore Family School of Electrical and Computer Engineering, Purdue University, West Lafayette,
IN, 47907.\protect\\
Corresponding author: M.~P.~E.~Apolinario, mapolina@purdue.edu}
\thanks{This work has been submitted to the IEEE for possible publication. Copyright may be transferred without notice, after which this version may no longer be accessible.}}

%
%

\markboth{}%
{}
%



\IEEEtitleabstractindextext{%
\begin{abstract} 
Spiking Neural Networks (SNNs) are bio-plausible models that hold great potential for realizing energy-efficient implementations of sequential tasks on resource-constrained edge devices. 
However, commercial edge platforms based on standard GPUs are not optimized to deploy SNNs, resulting in high energy and latency. 
While analog In-Memory Computing (IMC) platforms can serve as energy-efficient inference engines, they are accursed by the immense energy, latency, and area requirements of high-precision ADCs (HP-ADC), overshadowing the benefits of in-memory computations.
We propose a hardware/software co-design methodology to deploy SNNs into an ADC-Less IMC architecture using sense-amplifiers as 1-bit ADCs replacing conventional HP-ADCs and alleviating the above issues. 
Our proposed framework incurs minimal accuracy degradation by performing hardware-aware training and is able to scale beyond simple image classification tasks to more complex sequential regression tasks.
Experiments on complex tasks of optical flow estimation and gesture recognition show that progressively increasing the hardware awareness during SNN training allows the model to adapt and learn the errors due to the non-idealities associated with ADC-Less IMC.
Also, the proposed ADC-Less IMC offers significant energy and latency improvements, $2-7\times$ and $8.9-24.6\times$, respectively, depending on the SNN model and the workload, compared to HP-ADC IMC.

\end{abstract}

}

\maketitle

\IEEEdisplaynontitleabstractindextext

%
\IEEEpeerreviewmaketitle

\IEEEraisesectionheading{\section{Introduction}\label{sec:introduction}}

%
%
%
%
\IEEEPARstart{P}{rogress} in deep learning has allowed outstanding results in many cognitive tasks, albeit at the cost of increased model size and complexity \cite{Brown2020LanguageLearners, Dosovitskiy2021AnScale}.
Although such models are suitable for cloud-based systems with vast computational resources, they are not ideal for real-time edge applications, like autonomous drone navigation, where both low energy consumption and latency are crucial.
In recent years, Spiking Neural Networks (SNNs) have emerged as bio-plausible alternative to standard deep learning leading to energy-efficient models for sequential tasks. Their intrinsic features, such as inherent recurrence via membrane potential accumulation, sparsity, binary activations, event-based and spatio-temporal processing make them suitable for edge implementations.
A considerable amount of work has been done on training deep SNNs for image classification tasks, achieving accuracy comparable to those obtained by conventional Artificial Neural Networks (ANNs) \cite{Rueckauer2017ConversionClassification, Rathi2021DIET-SNN:Optimization, Fang2021DeepNetworks}. 
While image classification only demands spatial processing, SNNs also offer temporal processing capabilities, thanks to the inherent recurrence due to their membrane potential that serves as a memory.
Recent research has shown that SNNs can handle sequential tasks with performance at par with complex Gated Recurrent Unit (GRU) and Long Short-Term Memory (LSTM) based models\hl{, but} with much lesser parameters \cite{Cramer2022TheNetworks, Agrawal2021, Ponghiran2022SpikingLearning}.
We believe that sequential tasks, such as gesture recognition and event-based optical flow estimation are examples where all features of SNNs can be exploited to achieve better efficiency than the corresponding ANN implementations \cite{Lee2020Spike-FlowNet:Networks, Hagenaars2021Self-SupervisedNetworks}.

SNN deployment into existing commercial GPU-based edge hardware platforms are highly inefficient. This originates \hl{because the} present day Graphical Processing Units (GPUs) lack\hl{ } the ability to seamlessly process binary, sparse, and spatio-temporal inputs \cite{Turner2022MlGeNN:Networks}.
This has triggered the interest in domain-specific accelerators that can fully extract the benefits of SNNs \cite{Davies2018Loihi:Learning,  Merolla2014AInterface, Agrawal2021}.
Among the alternatives, In-Memory Computing (IMC) architectures turn out to be promising inference engines for deploying SNN models at a lower energy overhead and latency. This is possible because IMC architectures perform matrix-vector multiplications (MVM) within the memory crossbar arrays with a very high efficiency \cite{Roy2020In-MemoryOverview, Yu2020Compute-in-MemoryProspects}. 
Nevertheless, in conventional IMC architectures, the efficiency of MVMs is overshadowed by the high energy consumption of peripheral circuits used as interfaces between the analog and digital portions of the system. 
Specifically, High-Precision ADCs (HP-ADC) consume more than 60\% of the total system energy and occupy more than 81\% of the chip area, representing the major bottleneck in IMC architectures \cite{Roy2020In-MemoryOverview, Yu2020Compute-in-MemoryProspects, Christensen20222022Engineering}.

There have been several proposals to overcome the limitations imposed by HP-ADCs in \hl{analog} IMC architectures.
Some have focused on reducing ADC precision while compensating for accuracy degradation by changing IMC structures and splitting MVM operands \cite{Xia2016SwitchedNetwork, Kim2018Input-splittingArray}.
Later works focused on quantization-aware approaches to further reduce the ADC precision down to 1-bit \cite{Wei2020AComputing-in-Memory, Kim2021MappingADCs, Saxena2022TowardsLearning}.
Most of the above approaches focused on hardware optimization, either through performing fully analog operations or optimizing the design of ADCs \cite{Jiang2022ASub-arrays, Cao2021Neural-PIM:Peripherals} and focused only on ANNs for image classification tasks on simple datasets like MNIST and CIFAR10.
In this work, we propose a hardware/software co-design methodology to deploy SNNs into an ADC-Less IMC architecture by replacing HP-ADCs with sense-amplifiers as 1-bit ADCs (ADC-Less). We employ 1-bit partial sum quantization for the sense amplifiers and propose a  three-step hardware-aware training methodology to recover performance degradation. 

We performed experiments on image classification datasets such as CIFAR10 and CIFAR100 and on more complex vision tasks such as gesture recognition on the DVS128 gesture dataset \cite{Amir2017ASystem}, and optical flow estimation on the MVSEC dataset \cite{Zhu2018ThePerception}.
Our models demonstrate competitive application accuracy for all the tasks while achieving significant energy and latency improvements compared to HP-ADC IMC architectures and commercial GPU-based platforms, respectively.

The main contributions of the work can be summarized as follows. 

\begin{itemize}[\IEEEsetlabelwidth{Z}]
    \item An ADC-Less in-memory computing architecture for spiking neural networks with very low latency and energy consumption.
    \item An end-to-end hardware-aware training methodology for SNNs that compensates for the proposed architecture's quantization error (weights and binary partial sums). 
    \item Experiments validating that the proposed HW/SW co-design method is suitable for deploying spiking neural networks for tasks beyond image classification, such as optical flow estimation and gesture recognition.
    \item Generalization of the proposed scheme for different network architectures, such as plain CNNs, ResNets, and U-Net architectures.  
\end{itemize}

\section{Background}\label{sec:background}
 To bring energy-efficient machine intelligence to the edge, it is imperative that we effectively co-design the network and hardware architectures. In this section, we describe the SNN dynamics and the IMC hardware used to implement corresponding SNNs.
    
    
    \subsection{Spiking Neural Networks}\label{sec:snn}
   
    
    \begin{figure}[!t]
    \centering
    \includegraphics[width=\columnwidth]{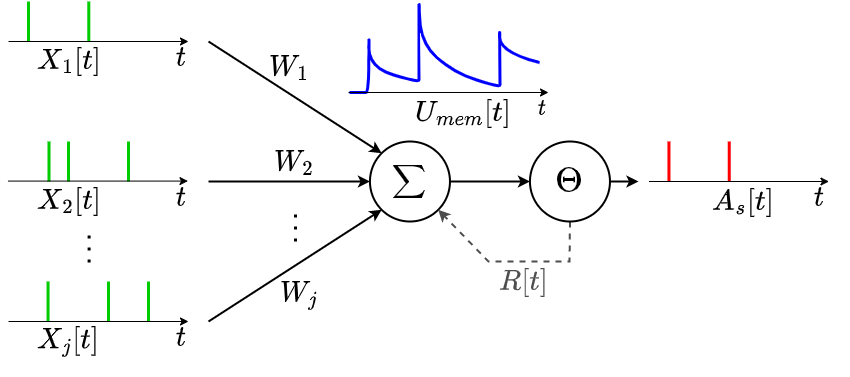}
    \caption{Dynamics of a LIF spiking neuron where the membrane potential, $U_{mem}$, integrates the incoming spikes, $X_j$, modulated by the weights,$W_j$, and leaks with a time constant, $\lambda$. When $U_{mem}$ reaches a threshold, $V_{th}$, the neuron fires generating an output spike, $A_s$. After the neuron fires, the membrane potential is reset.}
    \label{fig:spiking_neural_net}
    \end{figure}
    
    Among all the spiking models, the leaky-integrate-and-fire (LIF) neuron model is one of the simplest ones that retain the main features of biological neurons with binary activation (spikes). The LIF model can be described as follows:
    
    \begin{equation}
    \label{eqn:lif_model_acc}
    U_{mem}[t] = \lambda U_{mem}[t-1] + \sum\limits_{j} W_j X_{j}[t] - R[t-1]
    \end{equation}
    \begin{equation}
    \label{eqn:lif_model_fire}
    A_s[t] = \Theta (U_{mem}[t] - V_{th})
    \end{equation}
    \begin{equation}
    \label{eqn:lif_reset}
        R[t] =
        \left\{
        	\begin{array}{ll}
        		\lambda U_{mem}[t]A_s[t] & \mbox{, hard reset}\\
        		V_{th} A_s[t] & \mbox{, soft reset}
        	\end{array}
        \right.
    \end{equation}

    Where $\Theta$ is the Heaviside function, and $R[t]$ is a reset mechanism triggered when the neuron fires.
    For $R[t]$, there are two options: a hard reset and a soft reset. 
    The former sets $U_{mem}[t]$ to zero, while the latter reduces $U_{mem}[t]$ by a value equal to $V_{th}$.
    
    A visual representation of the spiking neural dynamics is shown in Fig.~\ref{fig:spiking_neural_net}. 
    The membrane potential accumulates inputs over time, where the inputs, $X_j[t]$, are multi-time-step binary spikes.
    The inherent recurrence of spiking models is due to membrane potential, $U_{mem}$, the hidden state evolving over time that can be described as short-term memory.
    Similarly, the leak parameter, $\lambda$, introduces a forgetting mechanism, similar in function to that of a "forget gate" in an LSTM model \cite{Ponghiran2022SpikingLearning}.
    These features enable SNNs to efficiently handle sequential tasks.
    

    \subsection{Crossbar-based In-memory Computing}\label{sec:crossbar_imc}

\begin{figure}[!t]
    \centering
    \includegraphics[width=\columnwidth]{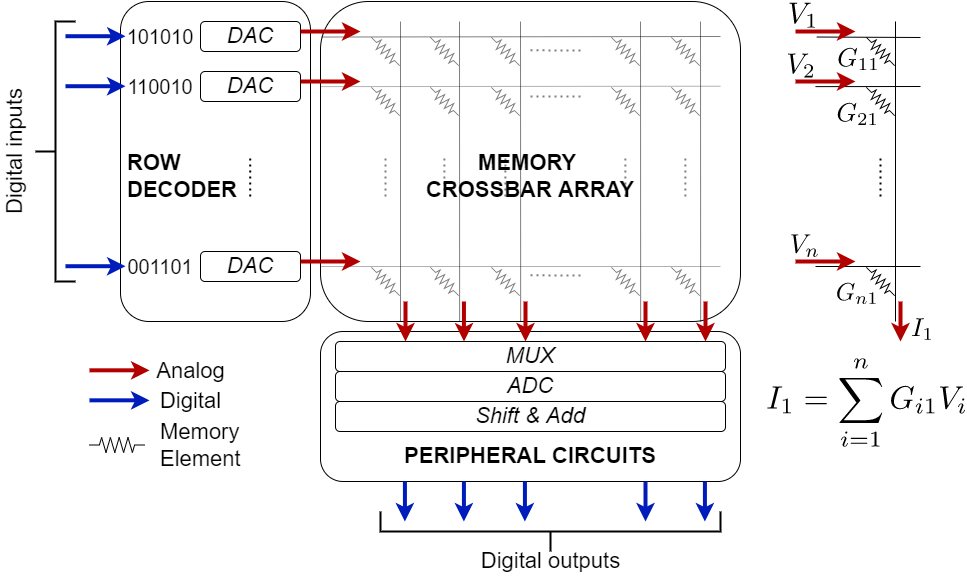}
    \caption{General scheme of a conventional crossbar-based analog MVM accelerator, showing the peripheral circuits.}
    \label{fig:imc_classic}
    \end{figure}

    
    In-memory computing (IMC) can be used to perform analog matrix-vector multiplication (MVM) operations with high energy efficiency. 
    The memory elements can be placed in a crossbar array, as shown in Fig.~\ref{fig:imc_classic}. 
    This configuration takes advantage of Kirchhoff's law to perform multiplication and accumulation (MAC) operations.
    The voltage  ($V_i$) on the input rows, also called wordlines, is multiplied by the conductances ($G_{i1}$) of the memory elements and accumulated on the current output ($I_1=\sum G_{i1}V_1$) of each column, called bitline. 
    This operation resembles the MAC operations performed by a neural network (NN), where the voltages are the NN's inputs, conductances are the weights of the NN, and the bitline current corresponds to the MVM operation output. 
    For \hl{an} SNN, the inputs are multi-time-step binary spikes (1/0). 
    SNN execution on IMC hardware involves MVM operation between the weight matrix and the binary input vector repeated for each time-step. 
    During each time step, MVM operation output is accumulated with the membrane potential and passed on to the neuron unit for application of activation function (LIF). 

    In order to map an SNN model into an IMC crossbar, the model parameters need to be quantized.
    Hence, the weights must have a finite resolution: number of bits per weight ($nb_W$).
    The number of memory cells used to represent the weights is determined by the bit-slicing weight resolution ($sb_W$), which is equal to the bit-cell resolution of the memory elements (note, memory elements can be multi-bit memories such as ReRAMs or PCRAMs, or single-bit such as SRAMs).
    Hence, the number of memory elements used per weight value is $nb_W/sb_W$. 
    The binary input activations in SNNs are applied through a 1-bit digital-to-analog converter (DAC). 
    At a particular timestep, multiple wordline are activated for matrix vector multiplication, which manifests itself as a current developed over the bitline. 
    The bitline current is passed through a multiplexer (MUX)
    connected to an analog-digital converter (ADC) which converts the analog current into a digital value.
    The multiplexer allows multiple bitlines to share the same ADC since it is impractical to have one high-precision ADC per bitline due to its large area.
    Ideally, the ADC requires a precision equal to $log_2(N_{WL})+sb_W-1$ where $N_{WL}$ is the number of wordlines used - typically equal to the crossbar size ($xbar$).
    Finally, the digitized values are shifted and added according to their MSB and LSB (most and least significant bits) positions. 
    The number of shifted operations is proportional to $(nb_W/sb_W)$.
    
     \subsection{\hl{Mapping SNN into crossbar arrays}}\label{sec:crossbar_and_snn}

    \begin{figure}[!t]
    \centering
    \includegraphics[width=0.9\columnwidth]{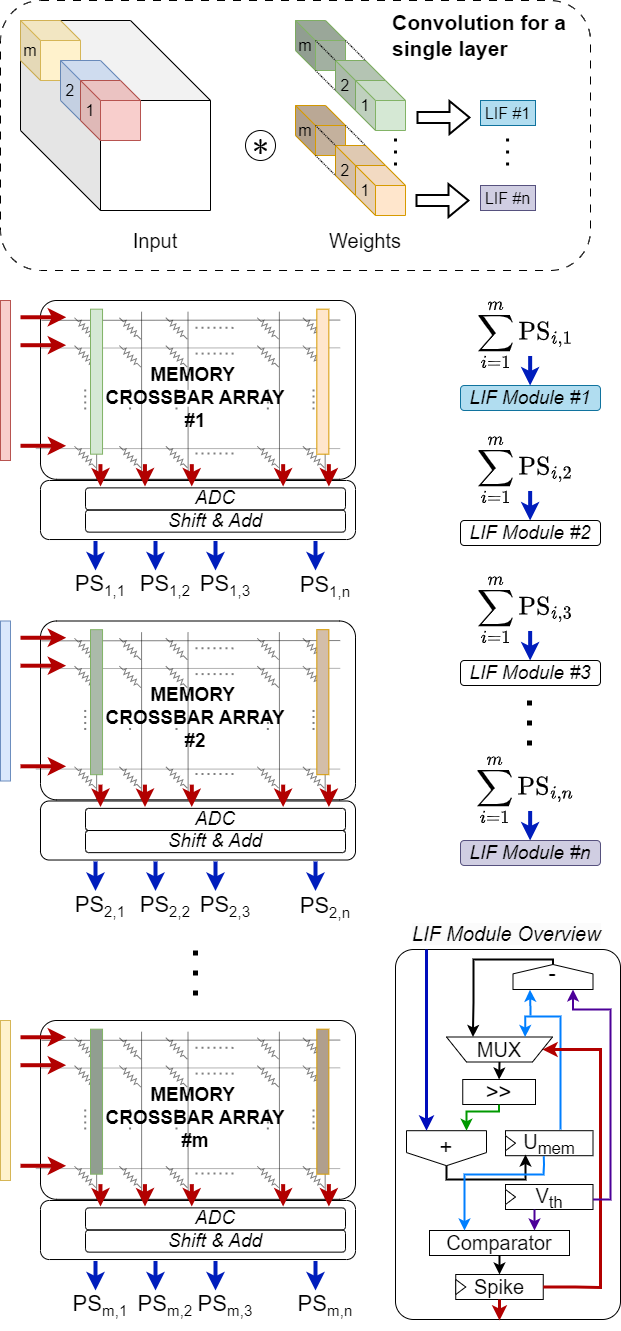}
    \caption{\hl{Diagram showing the partial-sum ($\text{PS}_{\text{i,j}}$) being produced when the synaptic weights of one layer with ``n" output neurons are split and mapped into ``m"  small crossbar arrays, where ``i" and ``j" are the indexes for the crossbar and neuron respectively. We also show the digital LIF neuron module implemented to perform spiking neural dynamics such as leakage, accumulation, spike generation, and reset.}}
    \label{fig:digital_lif}
    \end{figure}

    For a deep SNN, the size of layer weights is too large to be mapped into a single crossbar. 
    Hence, it has to be \hl{split and mapped into multiple crossbar arrays so that the number of rows multiplied by the number of crossbar arrays matches the number of weights per output channel.
    Then,} each array produces a partial-sum (PS) that \hl{has to be accumulated to obtain the final sum (as shown in Fig.~}\ref{fig:digital_lif}\hl{). 
    The final sum is then integrated into the membrane potential of the spiking neuron to produce a spike}. 
    Here, the ADC's resolution determines the accuracy of the PS. 
    Therefore, a high-precision ADC (HP-ADC) is required.
    However, HP-ADCs are expensive, both in terms of the chip area, and energy consumption \cite{Roy2020In-MemoryOverview, Christensen20222022Engineering}.
    Therefore, to optimize area, the ADC units have to be shared among multiple columns of the crossbar, which adds extra latency to the total operation of the crossbars. It is clear that the cost of the ADCs \hl{, to a certain extent,}  overshadow the potential benefits of employing crossbars to perform fast and energy-efficient MVMs. \cite{Yu2020Compute-in-MemoryProspects, Peng2019DNN+NeuroSim:Technologies}.
    

    In the above sections, we discussed a general crossbar-based IMC, independent of the memory technology. 
    Note, however, both CMOS and non-volatile technologies can be used as memory elements for the crossbar array.
    For the rest of the paper, we employ ReRAM as an example technology to show the effectiveness of our approach.
    
\section{Related Works}\label{sec:related_works}
Past works such as  \cite{Roy2020In-MemoryOverview, Peng2019DNN+NeuroSim:Technologies, Christensen20222022Engineering, Yu2020Compute-in-MemoryProspects, Xia2016SwitchedNetwork, Kim2021MappingADCs, Saxena2022TowardsLearning, Jiang2022ASub-arrays, Wei2020AComputing-in-Memory, Kim2018Input-splittingArray} identify ADCs as the principal bottleneck towards fully leveraging the energy efficiency of crossbar-based IMC architectures.
In \cite{Xia2016SwitchedNetwork}, the authors proposed an offline methodology to manually adjust the weight distribution of different crossbars to achieve a 1-bit partial sum using a thresholding operation, which is then merged to generate a 1-bit output to be fed into the next layer. They were able to eliminate both ADC and DAC modules without a significant accuracy drop.
However, their methodology was impractical for deeper networks, as the method's complexity and errors increase as the network becomes larger.

The authors in \cite{Kim2018Input-splittingArray} proposed an input-splitting scheme to represent a large neural network as multiple smaller models.
This allowed for a reduction of the ADC precision from 8-bits to 4-bits without a significant accuracy drop. However, as shown in \cite{Yu2020Compute-in-MemoryProspects}, a 4-bit ADC still consumes \hl{ } much energy and area compared to crossbar arrays.

Authors in \cite{Wei2020AComputing-in-Memory} proposed an iterative quantization-aware training method to partially overcome some of the hardware limitations imposed by the ReRAM-based IMC accelerators. 
The method involves quantizing the inputs and weights to a fixed resolution and then retraining the model considering the features of the crossbar architecture, and finally introducing the effects of the ADC resolution in the partial sum. 
Based on this approach \cite{Wei2020AComputing-in-Memory} reduces the precision of the ADC while the neural model partially compensates for the loss of accuracy. 
However, for ADC resolution below 4-bit, this method suffers from severe accuracy degradation. 

On the other hand, authors in \cite{Kim2021MappingADCs} explored quantization-aware training on binary ResNets targeting IMC architectures with low ADC precision.
Their approach resulted in small accuracy drop when using 1-bit ADCs when merging the partial sum followed by a batch normalization layer. Nevertheless their approach was limited to binary ResNet architectures, which are still challenging to train with high accuracy.
On a similar line, researchers in \cite{Saxena2022TowardsLearning} proposed a hardware-software co-design approach to eliminate the ADC modules by using only the sense-amplifier (SA) modules as 1-bit ADC and mitigate the accuracy loss using a quantization-aware training.
However, this method introduced additional parameters to scale the binary partial-sum to increase the representation ability of the partial sums.
Also, \cite{Jiang2022ASub-arrays} proposed a fully analog IMC macro to compute MVM without using ADCs by encoding the inputs as pulse-width modulated (PWM) signals. The main constraint being that the ANN models were limited by the crossbar size and were restricted to only 2-bit weights. Note, it required contiguous layers to be mapped into contiguous arrays in hardware, limiting the chip architecture.

All the works discussed above were intended for only ANN models and were evaluated on image classification tasks\hl{ }. 
Our work aims to extend the application to SNN models and scale to complex sequential regression tasks while being energy efficient.

\section{ADC-Less IMC Architecture}\label{sec:adcless_architecture}
\begin{figure}[!t]
\centering
\includegraphics[width=\columnwidth]{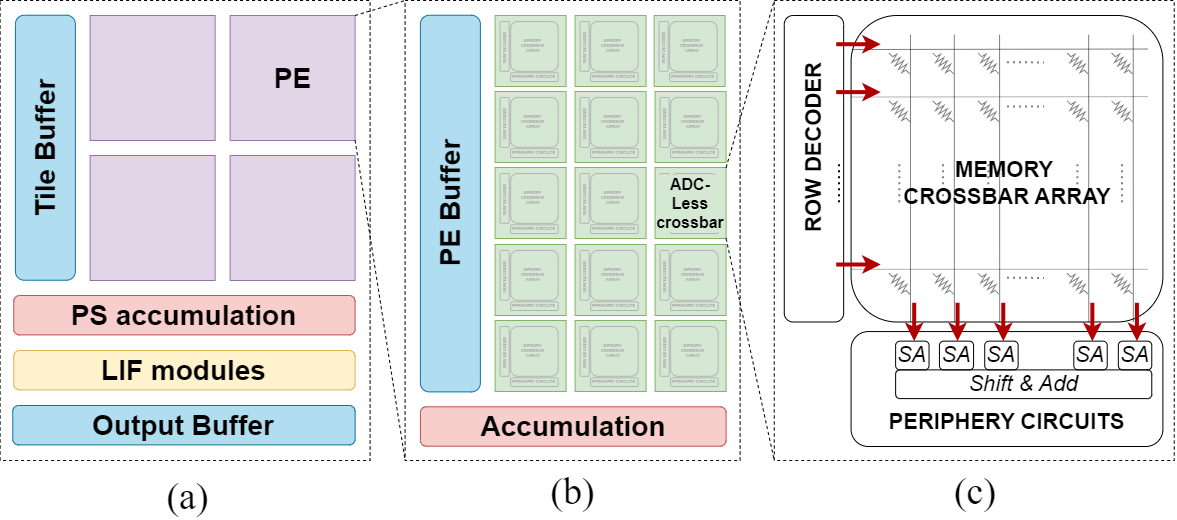}
\caption{\hl{Proposed scheme of the ADC-Less IMC for spiking neural networks. (a) Tile structure including LIF modules, (b) Processing Element (PE) structure, and (c) ADC-Less crossbar structure.}}
\label{fig:imc_adcless}
\end{figure}

In this section, we describe the proposed ADC-Less IMC architecture, which does not suffer from the performance bottleneck posed by ADCs in conventional IMC hardware design. 
\hl{For the general structure of the ADC-Less IMC, we adopt a conventional Tile-PE-Crossbar structure, similar to} \cite{Peng2019DNN+NeuroSim:Technologies},  as shown in Fig.~\ref{fig:imc_adcless}.
\hl{Two main differences in our design are the adoption of ADC-Less crossbars and computing LIF dynamics at the tile level.
First, in contrast to conventional HP-ADC crossbars} (Fig.~\ref{fig:imc_classic}), our ADC-Less \hl{crossbars} uses sense amplifier (SA) circuits at each bitline instead of a large high-precision ADC\hl{ shared by multiple bitlines}.
\hl{Section}~\ref{sec:adcless_crossbar} \hl{discusses the implications of this design choice at the crossbar level.
Second, including LIF modules at the Tile level, Fig.}~\ref{fig:imc_adcless}a\hl{ is done to allow sharing of the same resources to implement multiple spiking neurons.
Section}~\ref{sec:digital_lif} \hl{discusses the implementation and functionality of the LIF modules.}


    \subsection{\hl{ADC-Less Crossbars}}\label{sec:adcless_crossbar}
    \subsubsection{Binary partial-sum quantization}

    As described in Section~\ref{sec:crossbar_and_snn}, the bitline current of crossbar\hl{ arrays represent the analog value of the partial-sums (PS) in MVM operations}.
    \hl{Such analog} PS \hl{current} is \hl{then} digitized using an ADC, \hl{with} the resolution of the ADC \hl{determining }the dynamic range of PS (PS quantization). 
    For an SNN mapped into \hl{an HP-ADC }crossbar using binary memory cells ($sb_w = 1$), the ideal \hl{ADC resolution} required \hl{to preserve the dynamic range} is equal to $log_2(N_{WL})$. 
    \hl{In contrast, our proposed ADC-Less crossbars produce binary PS, using SA as 1-bit ADCs, Fig.~}\ref{fig:imc_adcless}.  
    \hl{This design decision is expected to significantly reduce energy and latency at the cost of introducing large quantization noise. 
    Such quantization effects and mitigation techniques are addressed later in} Section~\ref{sec:hardware_aware}.

    \subsubsection{\hl{Weight mapping}}\label{sec:mapping_weights}
    \hl{In addition to adopting binary PS, we evaluate two common schemes to map positive and negative 4-bit weights into binary memory-cell arrays and discuss how such schemes produce different PS}.
    \begin{figure}[!t]
    \centering
    \subfloat[]{\includegraphics[width=0.85\columnwidth]{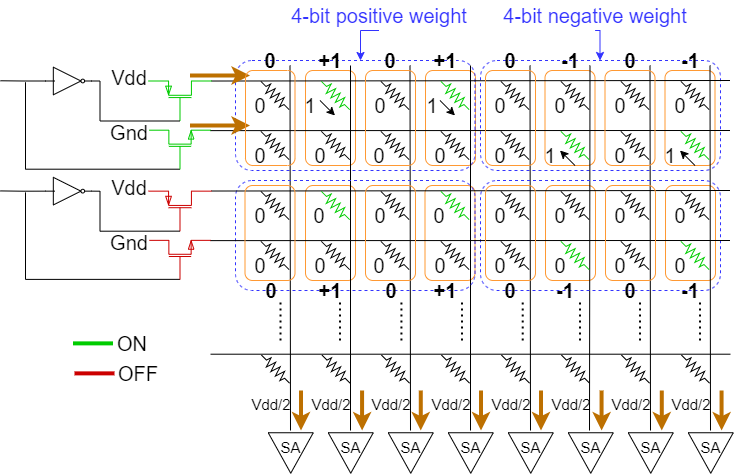}%
    \label{fig:mapping_1}}
    \hfil
    \subfloat[]{\includegraphics[width=0.8\columnwidth]{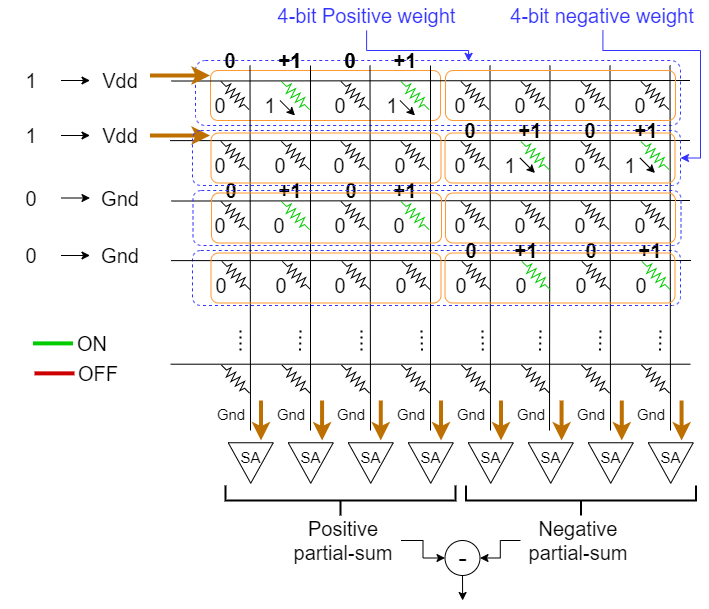}%
    \label{fig:mapping_2}}
    \caption{Two weight mapping schemes 
    (a) positive and negative weights sharing the same column with weights mapped in two rows, and (b) positive and negative weights in different columns with weights mapped in two columns.}
    \label{fig:mapping_scheme}
    \end{figure}
    
    The first mapping scheme is shown in Fig.~\ref{fig:mapping_1}, where both positive and negative weights are mapped into the same column using two rows\hl{, similar to }\cite{Saxena2022TowardsLearning}.
    The positive weights are mapped into the first row using an unsigned integer representation, while the second row is set to a high resistive state (Roff). 
    Similarly, the negative weights are mapped into the second row using an unsigned integer representation, while the first row is set to Roff.
    This strategy allows positive weights to charge the bitlines (BL) while negative ones to discharge them.
    For this mapping, the access to each row requires a pass transistor controlled by the binary input spikes.
    
    On the other hand, the second mapping scheme, shown in Fig.~\ref{fig:mapping_2}, uses different columns for positive and negative weights\hl{, similar to }\cite{Jiang2022ASub-arrays}. 
    The weights are mapped into one row using two consecutive 4-bit columns; the positive values are mapped into the first column, while the successive columns are set to Roff.
    Similarly, the negative values are mapped in the second column, while the first is set to Roff.
    In contrast to the first mapping scheme, the crossbar's wordlines can be accessed without additional circuitry. 
    
    The first mapping scheme quantizes the partial sum into $\pm1$ values for each column.
    In contrast, the second mapping scheme quantizes the partial sum for the positive and negative columns into 0 and 1 values that must be subtracted.

    \subsection{Digital LIF Neuron Module}\label{sec:digital_lif}
\begin{figure}[!t]
    \centering
    \includegraphics[width=\columnwidth]{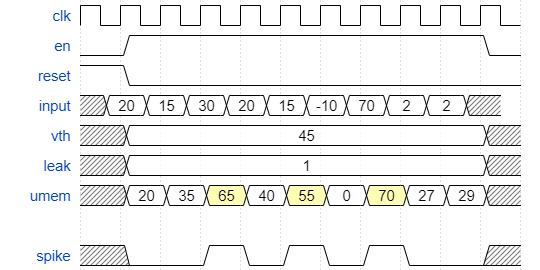}
    \caption{\hl{Simulated waveform of the digital LIF module during nine time-steps for a neuron with $Vth=45$ and $\lambda=1$
    . The values of the membrane potential (umem) that produce a spike are highlighted in yellow.}}
    \label{fig:lif_waveform}
    \end{figure}

    In addition to the ADC-Less IMC described in the previous section, we designed a digital module to perform the neural dynamics of the LIF model described in Section~\ref{sec:snn}. 
    Each LIF module\hl{, whose structure is shown in Fig.~}\ref{fig:digital_lif}, implements an individual neuron that accumulates the partial sum of multiple crossbars to a membrane potential ($U_{mem}$), storing it in a register, as described in (\ref{eqn:lif_model_acc}).
    In parallel, the additional circuits in the module monitor if the value of $U_{mem}$ is close to the threshold value ($V_{th}$) to fire an output spike, (\ref{eqn:lif_model_fire}), or start the reset mechanism, (\ref{eqn:lif_reset}).
    The module is implemented so that each time step of the LIF model corresponds to one clock cycle of the proposed architecture.
    \hl{A simulated waveform is shown in Fig.~}\ref{fig:lif_waveform}.

    \subsection{SNN inference with ADC-Less IMC}\label{sec:deploying_snn}
    During inference, the first layer of an SNN model receives an input sequence of frames, whose length (time-steps) depends on the workload ($5$ for optical flow, $10$ for image classification, and $20$ for gesture recognition).
    The output of this layer is a sequence of binary spikes with the same number of time-steps that pass through the rest of the model.
    In the following layers, the spikes are applied to the wordlines of the crossbars in a sequential manner, one spike at a time. 
    After the bitline current output is digitized according to the selected mapping scheme, described in Section~\ref{sec:mapping_weights}, the binary partial sum is shifted and accumulated appropriately.
    The partial sums from multiple crossbar arrays are added together and accumulated into the $U_{mem}$ register, as described in Section~\ref{sec:digital_lif}.
    Finally, the LIF module produces an output sequence of binary activation spikes for the next layer.

\section{Hardware-aware Training}\label{sec:hardware_aware}
The proposed ADC-Less architecture imposes some hardware constraints such as weight quantization (4-bit,$nb_w = 4$), memory cell resolution (1-bit per cell, $sb_w = 1$), and partial sum quantization (1-bit).
These features lead to a deviation from the ideal full-precision operation producing significant accuracy loss when a spiking model is deployed naively into the proposed architecture.

\begin{figure}[!t]
\centering
\includegraphics[width=\columnwidth]{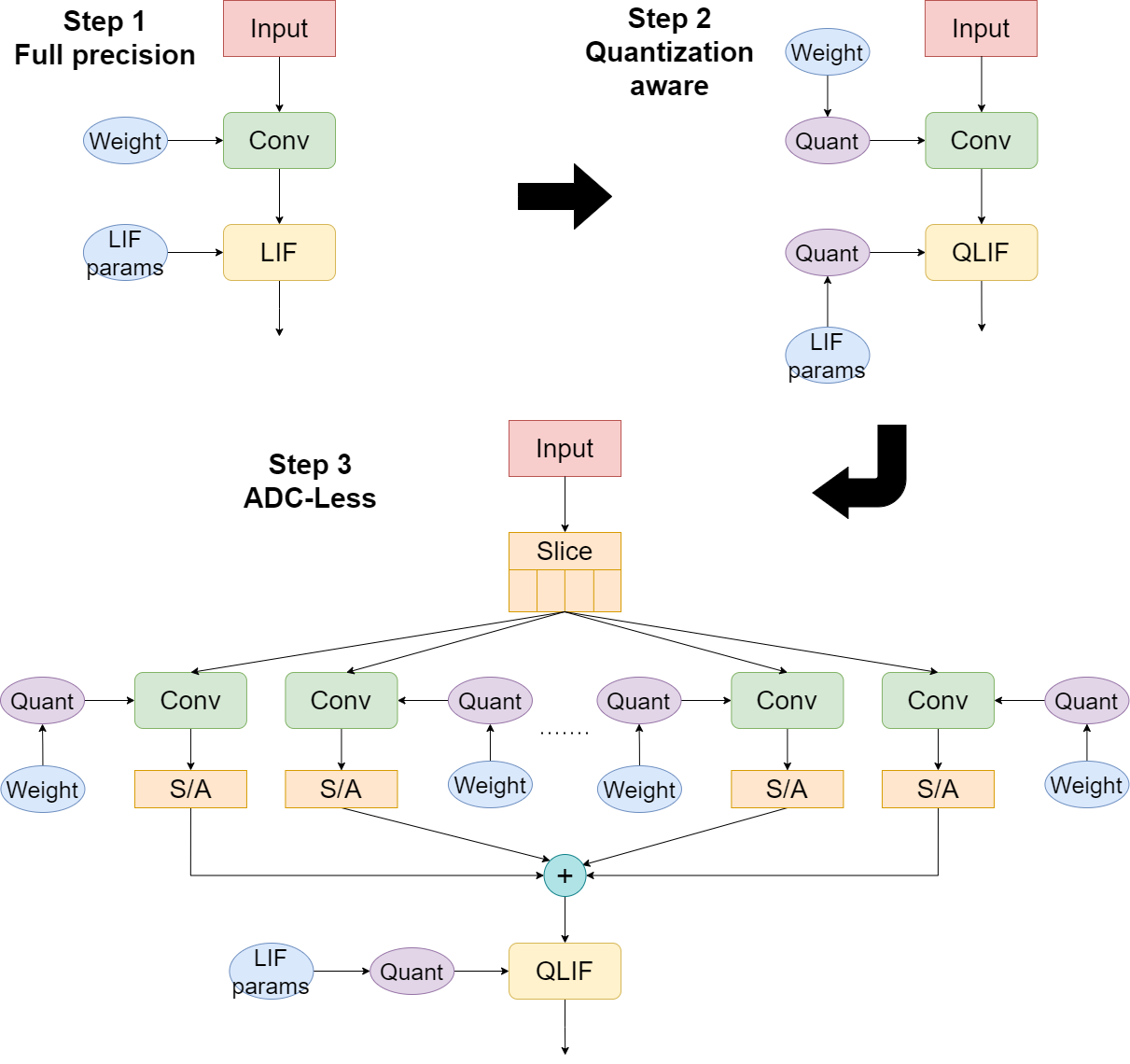}
\caption{The hardware-aware training methodology proposed is based on three consecutive training steps where the model resulting from the previous step is used as initialization for the next one.}
\label{fig:hardware_aware}
\end{figure}

Thus, we propose a three-step hardware-aware training to compensate for the deviation from an ideal inference.
The three-step method is shown in Fig.~\ref{fig:hardware_aware}, where the models are trained consecutively, increasing the hardware awareness in each step.
First, the method involves full-precision training using ANN-to-SNN conversion \cite{Rathi2021DIET-SNN:Optimization} or surrogate gradients \cite{Neftci2019SurrogateNetworks} for the spiking models.
In the second step, the model is re-trained using quantization-aware training with uniform 4-bit quantization.
Finally, in the third step, the denominated ``ADC-Less training'' fine-tunes the quantized model by exposing it to the weight mapping scheme, the memory-cell resolution, and the binary partial-sum quantization.

    \subsection{Full Precision Training}\label{sec:fp_snn}
    The first step of the proposed hardware-aware method begins with training a full-precision SNN model that would be used as model initialization for the next step. 
    For this purpose, we used two training approaches. 
    
    The first approach is based on ANN-to-SNN conversion followed by fine-tuning of the leak and threshold parameters as proposed in \cite{Rathi2021DIET-SNN:Optimization}, which has proven to be effective in training deep SNNs for image classification tasks with just a few time-steps.
    During the ANN-to-SNN conversion phase, the weights are copied to the SNN model. 
    At the same time, the threshold parameters ($V_{th}$) are set to the maximum activation value achieved during forward pass at each layer. This ensures that the spikes propagate through the entire mode without vanishing. Then, the leak ($\lambda$) and $V_{th}$ parameters are fine-tuned to achieve better accuracy.
    
    The second approach uses surrogate gradients \cite{Neftci2019SurrogateNetworks} to train the model from scratch. 
    One advantage of this approach is that it can be used for image classification and more complex sequential tasks, such as gesture recognition and optical flow estimation, where ANN-to-SNN conversion is unsuitable. 
    The surrogate gradient method approximates the derivative of the Heaviside function used by the LIF model during the backward pass of the training.

    \subsection{Quantization-aware Training}\label{sec:quantization_aware}
    During this step, we quantize the weights, LIF neuron membrane potential, the neuron threshold, and the leak to integer values to match hardware precision during inference. 
    The model is initialized from the previous step, and we perform quantization-aware training to fine-tune the model to integer-only data structures.  
    Our quantization scheme is designed to operate with SNNs, and it is based on \cite{Yao2021HAWQ-V3:Quantization}. 
    \hl{The quantization functions are described as follows:}

    \begin{equation}
        q_l = 2^{bits-1}-1
    \end{equation}
    \begin{equation}
        S_x(x) = \frac{max(|x|, 0)}{q_l}
        \label{eq:scale}
    \end{equation}
    \begin{equation}
        Q(x, S_x, q_l)=max(min(round(\frac{x}{S_x}), q_l), -q_l-1)
        \label{eq:q_def}
    \end{equation}
    \begin{equation}
        \frac{\partial Q(x, S_x, q_l)}{\partial x}=\frac{1}{S_x}
        \label{eq:grad_q}
    \end{equation}

    \hl{Where $q_l$ is the number of quantization levels corresponding to the bit resolution, $S_x$ is the scale of the parameter $x$, $Q$ is the function to quantize the parameter $x$ given a scale factor
    and the number of bits, and $\frac{\partial Q}{\partial x}$ is the surrogate gradient to be used during the backward pass.} 

    \begin{figure}[!t]
    \centering
    \includegraphics[width=\columnwidth]{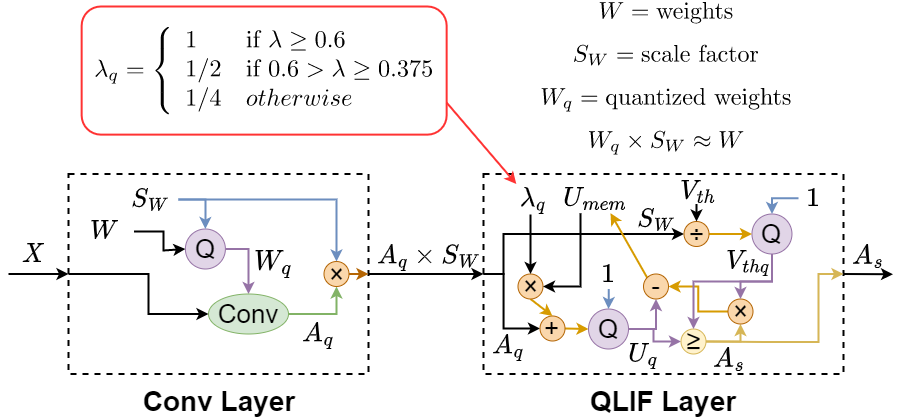}
    \caption{Quantization scheme for convolutional and LIF layers.}
    \label{fig:snn_quantization}
    \end{figure}

    The weights of convolutional and linear layers are symmetrically quantized to 4-bits. \hl{Hence, the weights ($W$) are separated into quantized weights ($W_q$) and a scale factor ($S_W$), using }(\ref{eq:scale}) and (\ref{eq:q_def}) as shown in Fig.~\ref{fig:snn_quantization}.
    
    The main difference from \cite{Yao2021HAWQ-V3:Quantization} is that we take advantage of the binary spikes to avoid an additional quantization of activations. 
    This reduces additional computations and error associated with activation quantization.
    Moreover, in \cite{Yao2021HAWQ-V3:Quantization}, the weights and activation of one layer are scaled based on its local parameters and the scale factors of previous layers.
    In contrast, we scale weights using only local parameters of that layer, so each layer is independently quantized.
    
    \hl{Also, the spiking states and parameters}, such as membrane potential and threshold, were scaled using the $S_W$ of the previous layer and quantized with $12$ bits \hl{using (}\ref{eq:q_def}). 
    In contrast, the leak parameter was quantized to powers of 2 ($2^{-n}$), with $n\in[0, 2]$, so that it could be implemented as a shift operation in hardware\hl{, as shown in Fig.~}\ref{fig:digital_lif}. 
    The \hl{complete} quantization process is shown in Fig.~\ref{fig:snn_quantization}.
    To avoid zero gradients produced by quantization operations, we use surrogate gradients to approximate the full precision gradient during the backward pass \hl{based on (}\ref{eq:grad_q}).


    \subsection{ADC-Less Training}\label{sec:adcless_aware}

    \begin{figure}[!t]
        \begin{algorithmic}[1]
        \REQUIRE $W_q$, $X$, $xbar$, $nb_W=4$, $sb_W=1$
        \ENSURE $A_q$
        \STATE $in\_ch \leftarrow W_q.shape()[1]$ \COMMENT{input channels}
        \STATE $k_H, k_W \leftarrow W_q.shape()[2:]$ \COMMENT{kernel size}
        \STATE $n\_groups \leftarrow ceil(\frac{in\_ch\times k_H \times k_W}{xbar})$

        \WHILE{$in\_ch \% n\_groups == 0$}
            \STATE $n\_groups \leftarrow n\_groups + 1 $
        \ENDWHILE
        \STATE $W_{pos-q, nb_W} \leftarrow \textbf{binarize}(\textbf{relu}(W_q), nb_W, sb_W) $

        \STATE $W_{neg-q, nb_W} \leftarrow \textbf{binarize}(\textbf{relu}(-W_q), nb_W, sb_W) $

        \COMMENT{If using the first mapping scheme}
        \STATE $out\_acc \leftarrow 0$
        \STATE $W_{q, nb_W} \leftarrow W_{pos-q, nb_W} - W_{neg-q, nb_W}$
        \FOR{$i$ in \textbf{range}$(0, nb_W/sb_W)$}
            \STATE $w \leftarrow \textbf{cat}(\textbf{chunk}(W_{pos-q, i}, n\_groups, dim=1))$
            \STATE $out \leftarrow \textbf{conv}(w, X, groups=n\_groups)$
            \STATE $out \leftarrow \textbf{adc\_less\_act}(out)$ \COMMENT{$\pm 1$ values}
            \STATE $out\_acc \leftarrow out\_acc + 2^i\times out$
        \ENDFOR
        \STATE $A_{acc} \leftarrow out\_acc$
        
        \COMMENT{End of first mapping scheme}
        
        \COMMENT{If using the second mapping scheme}
        \STATE $out\_pos \leftarrow 0$
        \FOR{$i$ in \textbf{range}$(0, nb_W/sb_W)$}
            \STATE $w \leftarrow \textbf{cat}(\textbf{chunk}(W_{pos-q, i}, n\_groups, dim=1))$
            \STATE $out \leftarrow \textbf{conv}(w, X, groups=n\_groups)$
            \STATE $out \leftarrow \textbf{adc\_less\_act}(out)$ \COMMENT{$0, 1$ values}
            \STATE $out\_pos \leftarrow out\_pos + 2^i\times out$
        \ENDFOR

        \STATE $out\_neg \leftarrow 0$
        \FOR{$i$ in \textbf{range}$(0, nb_W/sb_W)$}
            \STATE $w \leftarrow \textbf{cat}(\textbf{chunk}(W_{neg-q, i}, n\_groups, dim=1))$
            \STATE $out \leftarrow \textbf{conv}(w, X, groups=n\_groups)$
            \STATE $out \leftarrow \textbf{adc\_less\_act}(out)$ \COMMENT{$0, 1$ values}
            \STATE $out\_neg \leftarrow out\_neg + 2^i\times out$
        \ENDFOR
        \STATE $A_{acc} \leftarrow out\_pos - out\_neg$
        
        \COMMENT{End of second mapping scheme}
        
        \STATE $A_{acc,g} \leftarrow \textbf{chunk}(A_{acc}, n\_groups, dim=1)$
        \STATE $A_q \leftarrow 0$
        \FOR{$j$ in \textbf{range}$(0, n\_groups)$}
            \STATE $A_q \leftarrow A_q + A_{acc,j}$
        \ENDFOR

        \RETURN $A_q$
        \end{algorithmic}
    \caption{Pseudo-code for ADC-Less convolution operation.}
    \label{fig:adc_less_algorithm}
    \end{figure}

    After the quantization-aware training, the model \hl{is} fine-tuned, considering the weight mapping scheme ($nb_W$ weight bit resolution and $sb_W$ bit slicing), the crossbar array size ($xbar$), and the partial-sum quantization.
    Here, the conventional convolution operation (Conv), used in the quantization scheme shown in Fig.~\ref{fig:snn_quantization}, was replaced with the ADC-Less convolution described in Fig.~\ref{fig:adc_less_algorithm}.
    This new convolution was implemented by \hl{separating the weights into their binary components [MSB ... LSB]} and splitting the input channel dimension of the kernel into a certain number of groups computed according to the $xbar$ and kernel sizes.
    The split kernel is used to perform a grouped convolution over the spike inputs ($X$), to obtain PSs.
    The PSs are then quantized according to the weight mapping scheme, producing binary PS that are then shifted and added properly.
    \hl{For the binary PS, we use the sign function,} 
    $$sign(x)=\begin{cases}
    1,& \text{if } x> 0\\
    -1,& \text{if } x< 0\\
    0,& \text{otherwise}
\end{cases}\text{,}$$ 
    \hl{and the Heaviside function,} $$h(x)=\begin{cases}
    1,& \text{if } x> 0\\
    0,& \text{otherwise}
\end{cases}\text{,}$$ \hl{for the first and second mapping schemes, respectively.}

    Finally, the PSs are accumulated into the final activation ($A_q$) which goes to the input of the QLIF layer to perform the spiking dynamics. 

    We also use surrogate gradients to avoid the \hl{zero} gradient problem due to the kernel slicing and binary partial sum. 
    The gradient is masked with the binary bits mapped into the crossbar arrays for the slicing.
    And for \hl{both} binary partial sum \hl{functions}, the gradient\hl{s are} approximated by $$g(x)=\frac{1}{1+\alpha x^2}$$

\section{SNN Models}\label{sec:snn_models}
This section describes four SNN models used in our experiments (two for image classification, one for gesture recognition, and one for optical flow).

\begin{figure}[!t]
\centering
\subfloat[]{\includegraphics[height=1.6in]{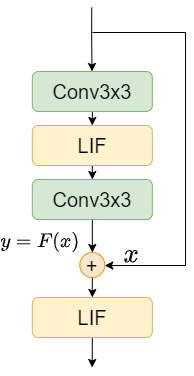}%
\label{fig:regular_residual}}
\hfil
\subfloat[]{\includegraphics[height=1.6in]{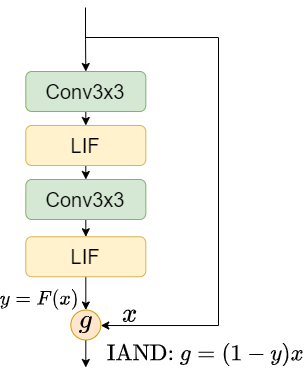}%
\label{fig:sew_block}}
\caption{Residual blocks for SNN: (a) regular residual block, and (b) IAND spike-element-wise (SEW) block.}
\label{fig:residual_block}
\end{figure}

As discussed earlier, this work exploits the binary activations in SNNs. 
For this reason, our SNN models were carefully designed to preserve the spikes at all layers during both training and inference stages.
For instance, all the models use max pooling instead of average pooling since the latter destroys spikes.
We modify the dropout layers to avoid the scaling factor used during training; initialize them at the beginning of a sequence, and keep them frozen until the end. 
The most crucial change was made to the residual layers used in three of the four SNN models.
A regular residual block, as shown in Fig.~\ref{fig:regular_residual}, merges the outputs from the direct path and skip-connection by adding both signals ($x+y$), and then recovers the spikes by using a LIF layer.
This kind of block works well for full precision models using ANN-to-SNN conversion methods \cite{Rathi2021DIET-SNN:Optimization}. However, it is not compatible with our quantization-aware training scheme that aims to perform only integer-integer computations.
For this reason, we adopt the IAND Spike-Element-Wise (SEW) block proposed in \cite{Fang2021DeepNetworks} that merges $x$ and $y$ signals by using a binary IAND logical operation (Fig.~\ref{fig:sew_block}).
This logical operation can be implemented much more efficiently on hardware compared to an adder by using just two logic gates. 
In addition, the SEW block has been proven to be suitable for training deep models with surrogate gradients \cite{Fang2021DeepNetworks}.

    \subsection{VGG16 and ResNet20}\label{sec:image_models}
    For image classification, we use spiking VGG16 and ResNet20 models.
    In order to ensure that all the layers receive only binary activations, we made minor changes to both architectures.
    For both models, we use only max pooling layers instead of average pooling to preserve the integrity of the spikes.
    Moreover, for the ResNet20, we use the SEW residual block shown in Fig.~\ref{fig:sew_block}, to ensure that all the inputs are binary.
    
    For quantization, we use 8-bit weights for the first and last layer with HP-ADC and 4-bits weights for the other layers with ADC-Less.

    \subsection{DVSNet}\label{sec:dvsnet}
    For gesture recognition, we designed a model called DVSNet based on the IAND SEW blocks described earlier.
    The input layer is a Conv3x3 with $32$ output channels, followed by five SEW blocks with $32$ input channels, where each block uses a MaxPool2x2 layer at the output.
    The output spikes of the convolutional layers are accumulated and flattened to be classified by a final Linear128x11 layer.
    
    For quantization, we use 4-bit weight for all the layers.
    The first and last layers use HP-ADC, and the other layers use ADC-Less.

    \subsection{Fully-spiking FlowNet (FSFN) model}\label{sec:fsfn}

    \begin{figure}[!t]
    \centering
    \includegraphics[width=\columnwidth]{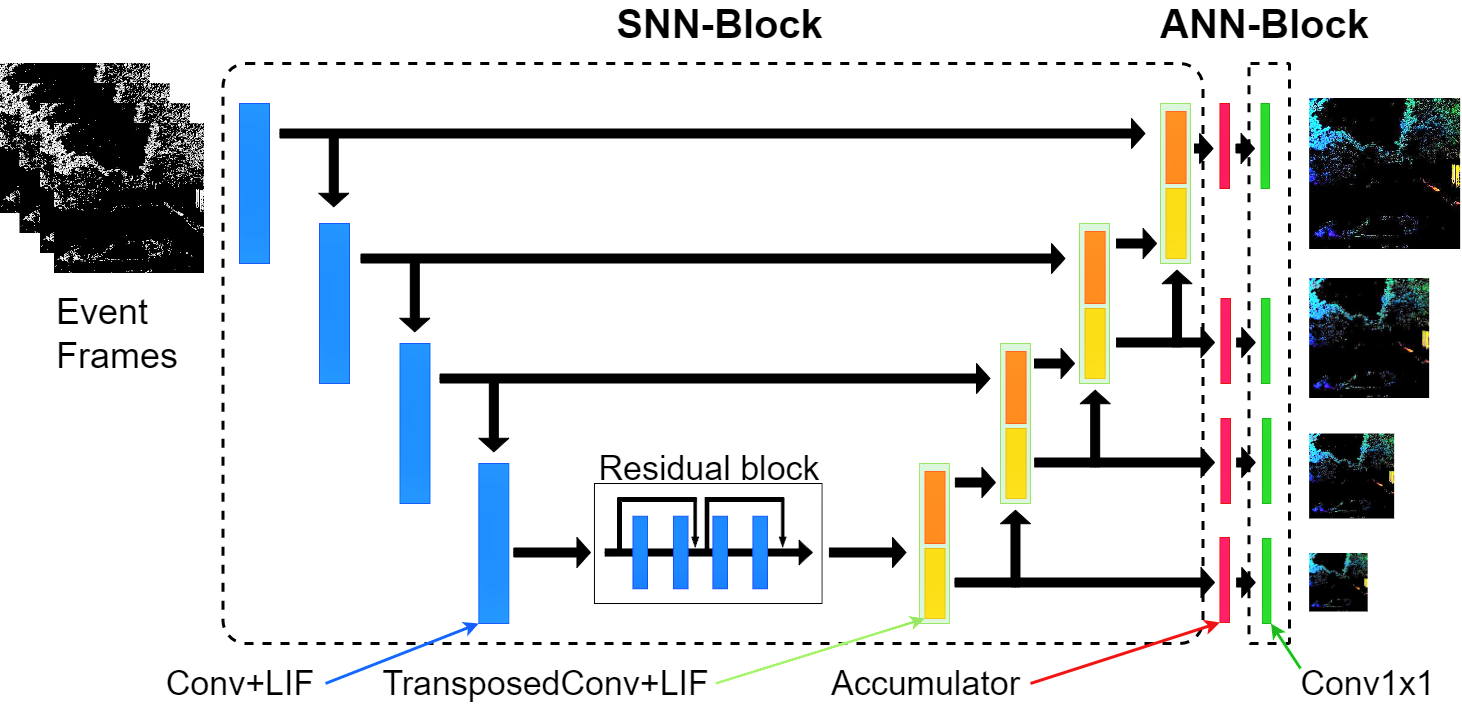}
    \caption{Fully-Spiking FlowNet (FSFN) encoder-decoder architecture. Both encoder and decoder are spiking layers, and only the last layer used for the final estimation of the optical flow is composed of Conv1x1 layers. During the training, outputs of all four levels are used to compute loss and guide the training, but only the output with the highest resolution is used during inference time.}
    \label{fig:spikeflownet}
    \end{figure}
    
    For optical flow estimation, we implemented a fully-spiking model based on the hybrid SNN-ANN Spike-FlowNet \cite{Lee2020Spike-FlowNet:Networks} model.
    We denominate our model as Fully-Spiking FlowNet (FSFN).
    FSFN has some minor changes with respect to the original Spike-FlowNet architecture in terms of kernel size, channels, and input sizes.
    These changes ensure that all layers receive only binary inputs.
    Since optical flow is an analog quantity, we maintain the last layer as an analog Conv1x1 layer (ANN block). This layer receives the accumulated outputs produced by the SNN block as inputs, as shown in Fig.~\ref{fig:spikeflownet}.
    
    Four Conv1x1 layers compound the ANN-block with $32$ channel inputs and two channel outputs.
    This block produces outputs at different scales that are helpful to guide the model during training, as they connect the global loss with the first layers allowing learning coarse and fine flow at multiple scales.
    However, only the output layer with the highest resolution is used during the inference stage.

    For quantization, we use the following configuration: 8-bit weights for the first layer with HP-ADC, 4-bits weights for the rest of the SNN-Block with ADC-Less, and full precision Conv1X1 layers for the ANN-Block. We use full precision in the ANN-Block because the optical flow prediction is a continuous value requiring high precision.

\section{Experimental Results}\label{sec:experimental_results}

    \subsection{Computational evaluation on different tasks}\label{sec:computational_evaluation}
    In this section, we evaluate the hardware-aware training discussed in Section~\ref{sec:hardware_aware} for multiple tasks such as image classification, gesture recognition, and optical flow estimation. In all the cases, we compare the ADC-Less model to a quantized model using HP-ADCs. 
    All models were selected using cross-validation with a three-partition dataset (train, validation, and test sets).
        \subsubsection{Experiments on Image Classification}\label{sec:image_classification}  
        Although image classification is not the most suitable computer vision task for SNNs, it helps to set a baseline for comparing our work with other proposals focused on ANNs.
        As discussed in Section~\ref{sec:related_works}, most previous works tested their proposed schemes on MNIST and CIFAR10. 
        However, the MNIST dataset is a simple image classification task, so we only focus on CIFAR10 dataset and extend our results to CIFAR100 dataset. 
        For both datasets, we use the following distribution: $50$k images for training, $5$k for validation, and $5$k for testing.
        
        Here, we evaluate the hardware-aware training on the two SNN models discussed in Section~\ref{sec:image_models}.
        The full-precision (FP) VGG16 was trained based on an ANN-SNN conversion + fine-tuning scheme \cite{Rathi2021DIET-SNN:Optimization} with 5~time-steps, while FP ResNet20 was trained from scratch using surrogate gradients with 10~time-steps.  
        
        After the FP training, we proceed with the quantization-aware training followed by the ADC-Less training.
        In both cases, the models were trained using $10$~time-steps. 
        The accuracy values obtained for the test set are shown in Table~\ref{table:image_classification}.
        We can see that quantization-aware training improves the accuracy of the models in all cases because quantization has a regularization effect that supports SNN learning.
        Then, for the ADC-Less aware training, we used the second mapping scheme, shown in Fig.~\ref{fig:mapping_2}.

        Table~\ref{table:image_classification} shows that the accuracy drop for both ADC-Less models (VGG16 and ResNet20) on CIFAR10 is around $1.16-2.42\%$ compared to the HP-ADC, depending on the size of the crossbar array. 
        This range is slightly higher than the accuracy drops reported in similar works focused on ANNs \cite{Saxena2022TowardsLearning, Kim2021MappingADCs}. 
        The accuracy drop is a natural effect of using spiking models for static tasks. 
        Note, we report accuracy values obtained for the test set in a cross-validation scheme, while previous works only reported accuracy on the validation set. 
        Similarly, for CIFAR100, VGG16 shows a drop in accuracy in the range of $0.23-1.78\%$ depending on the crossbar size.
        In all the cases, the ADC-Less training can minimize the accuracy drop achieving comparable results to the HP-ADC models.
        \hl{In addition, the average number of spikes per neuron and timesteps (Avg. spikes) remains relatively consistent across all models, as illustrated in Table 1.}

        \begin{equation}
        \text{Avg. spikes (\%)} = \frac{\text{\#fired spikes}\times 100\%}{\text{\#total neurons}\times\text{\#time-steps} }
        \end{equation}

        \begin{table}[!t]
        \renewcommand{\arraystretch}{1.3}
        \caption{Test accuracy of spiking models for image classification on CIFAR10 and CIFAR100 }
        \label{table:image_classification}
        \centering
        \begin{tabular}{|llllll|}
        \hline
        \multicolumn{1}{|c|}{\multirow{2}{*}{\textbf{Metrics}}} & \multicolumn{1}{c|}{\multirow{2}{*}{\textbf{FP}}} & \multicolumn{1}{c|}{\multirow{2}{*}{\textbf{HP-ADC}}} & \multicolumn{3}{c|}{\textbf{ADC-Less}} \\ \cline{4-6} 
        \multicolumn{1}{|c|}{} & \multicolumn{1}{c|}{} & \multicolumn{1}{c|}{} & \multicolumn{1}{c|}{\textbf{32}} & \multicolumn{1}{c|}{\textbf{64}} & \multicolumn{1}{c|}{\textbf128} \\ \hline
        \multicolumn{6}{|c|}{CIFAR10 (VGG16)} \\ \hline
        \multicolumn{1}{|l|}{Acc@1}            & \multicolumn{1}{l|}{88.89}        & \multicolumn{1}{l|}{91.93}           & \multicolumn{1}{l|}{90.70}             & \multicolumn{1}{l|}{90.44}             & 89.52                                  \\ \hline
        \multicolumn{1}{|l|}{Gap}              & \multicolumn{1}{l|}{-}            & \multicolumn{1}{l|}{-}               & \multicolumn{1}{l|}{-1.23}             & \multicolumn{1}{l|}{-1.49}             & -2.41                                  \\ \hline
        \multicolumn{1}{|l|}{\hl{Avg. spikes (\%)}}              & \multicolumn{1}{l|}{\hl{7.37}}            & \multicolumn{1}{l|}{\hl{5.88}}               & \multicolumn{1}{l|}{\hl{5.93}}             & \multicolumn{1}{l|}{\hl{6.47}}             & \hl{6.99}                                  \\ \hline
        \multicolumn{6}{|c|}{CIFAR10 (ResNet20)}                                                                                                                                                                                                     \\ \hline
        \multicolumn{1}{|l|}{Acc@1}            & \multicolumn{1}{l|}{89.97}        & \multicolumn{1}{l|}{90.12}           & \multicolumn{1}{l|}{88.96}             & \multicolumn{1}{l|}{88.46}             & 87.70                                  \\ \hline
        \multicolumn{1}{|l|}{Gap}              & \multicolumn{1}{l|}{-}            & \multicolumn{1}{l|}{-}               & \multicolumn{1}{l|}{-1.16}             & \multicolumn{1}{l|}{-1.66}             & -2.42                                  \\ \hline
        \multicolumn{1}{|l|}{\hl{Avg. spikes (\%)}}              & \multicolumn{1}{l|}{\hl{7.52}}            & \multicolumn{1}{l|}{\hl{10.36}}               & \multicolumn{1}{l|}{\hl{8.38}}             & \multicolumn{1}{l|}{\hl{8.02}}             & \hl{7.50}                                  \\ \hline
        \multicolumn{6}{|c|}{CIFAR100 (VGG16)}                                                                                                                                                                                                       \\ \hline
        \multicolumn{1}{|l|}{Acc@1}            & \multicolumn{1}{l|}{66.39}        & \multicolumn{1}{l|}{68.71}           & \multicolumn{1}{l|}{68.48}             & \multicolumn{1}{l|}{67.11}             & 66.93                                  \\ \hline
        \multicolumn{1}{|l|}{Gap}              & \multicolumn{1}{l|}{-}            & \multicolumn{1}{l|}{-}               & \multicolumn{1}{l|}{-0.23}             & \multicolumn{1}{l|}{-1.6}              & -1.78                                  \\ \hline
        \multicolumn{1}{|l|}{\hl{Avg. spikes (\%)}}              & \multicolumn{1}{l|}{\hl{7.62}}            & \multicolumn{1}{l|}{\hl{6.89}}               & \multicolumn{1}{l|}{\hl{5.90}}             & \multicolumn{1}{l|}{\hl{5.94}}             & \hl{5.78}                                  \\ \hline
        \end{tabular}
        \end{table}

        \subsubsection{Experiments on Gesture Recognition}\label{sec:gesture_recognition}
        As discussed in Section~\ref{sec:snn}, SNNs are able to handle spatio-temporal data, such as those produced by event-based cameras.
        Event-based cameras detect log-scale brightness changes asynchronously and independently on each pixel-array element, producing positive and negative binary impulse trains for the corresponding pixel \cite{Lichtsteiner2008ASensor}. 
        This results in binary, sparse, and high-temporal resolution representations, features that make them a good fit for SNNs.
        
        Here, we test DVSNet, described in Section~\ref{sec:snn_models}, for gesture recognition on the DVS128~Gesture~dataset~\cite{Amir2017ASystem}. This dataset contains event data recorded from 29 subjects performing 11 hand gestures in 3 different lighting conditions.
        The data is augmented by slicing the original sequences into time windows of $1.5$ seconds without overlapping. 
        Then, event frames are generated by accumulating the events during windows of $75~ms$ ($20$~time-steps per sequence) and finally downsamples to a $64\times64$ size. 
        This preprocessing results in ~$4$k sequences for training, ~$500$ for validation, and ~$500$ for testing.
        
        We train our DVSNet model from scratch using surrogate gradients following the proposed three-step hardware-aware training.
        For the quantization-aware training, we use a 4-bit weight quantization for all the layers.
        Then, for the ADC-Less training, we use the second weight mapping scheme discussed in Section~\ref{sec:adcless_aware} since it encourages sparsity. 
        Similar to the previous section, we evaluate the ADC-Less training with three crossbar array sizes, $32$, $64$, and $128$.
       
        \begin{table}[!t]
        \renewcommand{\arraystretch}{1.3}
        \caption{Test accuracy of DVSNet for gesture recognition on DVS128 Gesture dataset}
        \label{table:gesture_recognition}
        \centering
        \begin{tabular}{|l|l|l|lll|}
        \hline
        \multicolumn{1}{|c|}{\multirow{2}{*}{\textbf{Metrics}}} & \multicolumn{1}{c|}{\multirow{2}{*}{\textbf{FP}}} & \multicolumn{1}{c|}{\multirow{2}{*}{\textbf{HP-ADC}}} & \multicolumn{3}{c|}{\textbf{ADC-Less}} \\ \cline{4-6} 
        \multicolumn{1}{|c|}{} & \multicolumn{1}{c|}{} & \multicolumn{1}{c|}{} & \multicolumn{1}{c|}{\textbf{32}} & \multicolumn{1}{c|}{\textbf{64}} & \multicolumn{1}{c|}{\textbf128} \\ \hline
        Acc@1 & 94.56 & 94.95 & \multicolumn{1}{l|}{95.15} & \multicolumn{1}{l|}{93.40} & 91.85 \\ \hline
        Gap & - & - & \multicolumn{1}{l|}{+0.2} & \multicolumn{1}{l|}{-1.55} & -3.1 \\ \hline
        \hl{Avg. spikes (\%)} & \hl{5.01} & \hl{5.48} & \multicolumn{1}{l|}{\hl{5.67}} & \multicolumn{1}{l|}{\hl{6.56}} & \hl{6.26} \\ \hline
        
        \end{tabular}
        \end{table}

        The accuracy values measured on the test set are shown in Table~\ref{table:gesture_recognition}.
        It is worth noting that the ADC-Less model with a crossbar size of $32$ is slightly better than the HP-ADC model.
        It is because of the high sparsity in the intermediate layer activations that result in just a few wordlines being activated at the same time.
        On the other hand, when increasing the xbar size to $64$ and $128$ there is an accuracy drop of $1.55\%$ and $3.1\%$, respectively. 
        
        These results so far show that the hardware-aware training proposed can also be used for sequential classification tasks with no significant accuracy loss.

        \subsubsection{Experiments on Optical Flow Estimation}\label{sec:optical_flow}

        \begin{table*}[!t]
        \renewcommand{\arraystretch}{1.3}

        \caption{Comparison of the AEE metric on MVSEC \cite{Zhu2018ThePerception} dataset [AEE lower is better]}
        \label{table:mvsec_results}
        \centering
        \begin{tabular}{|l|l|l|l|l|l|l|l|}
        \hline
        \multicolumn{1}{|c|}{\textbf{Models}} & \multicolumn{1}{c|}{\textbf{Quant}} & \multicolumn{1}{c|}{\textbf{Type}} & \multicolumn{1}{c|}{\textbf{\begin{tabular}[c]{@{}c@{}}Outdoor\_day1\\ AEE\end{tabular}}} & \multicolumn{1}{c|}{\textbf{\begin{tabular}[c]{@{}c@{}}Indoor\_flying1\\ AEE\end{tabular}}} & \multicolumn{1}{c|}{\textbf{\begin{tabular}[c]{@{}c@{}}Indoor\_flying2\\ AEE\end{tabular}}} & \multicolumn{1}{c|}{\textbf{\begin{tabular}[c]{@{}c@{}}Indoor\_flying3\\ AEE\end{tabular}}}  & \multicolumn{1}{c|}{\textbf{\hl{Avg. spikes (\%)}}} \\ \hline
        FSFN\textsubscript{FP} (Ours) & No & Spiking & 0.51 & 0.82 & 1.21 & 1.07 & \hl{9.11} \\ \hline
        FSFN\textsubscript{HP-ADC} (Ours) & Yes & Spiking & 0.48 & 0.85 & 1.29 & 1.13 & \hl{11.61} \\ \hline
        FSFN\textsubscript{ADC-Less$\mid$32} (Ours) & Yes & Spiking & 0.65 & 0.91 & 1.41 & 1.18  & \hl{7.68} \\ \hline
        FSFN\textsubscript{ADC-Less$\mid$64} (Ours) & Yes & Spiking & 0.65 & 0.88 & 1.39 & 1.18 & \hl{8.14} \\ \hline
        FSFN\textsubscript{ADC-Less$\mid$128} (Ours) & Yes & Spiking & 0.65 & 0.94 & 1.54 & 1.30 & \hl{8.36} \\ \hline
        Spike-FlowNet \cite{Lee2020Spike-FlowNet:Networks} & No & Hybrid & 0.47 & 0.84 & 1.28 & 1.11 & \hl{-}  \\ \hline
        EV-FlowNet  \cite{Zhu2018EV-FlowNet:Cameras} & No & Analog & 0.49 & 1.03 & 1.72 & 1.53 & \hl{-}  \\ \hline
        Zhu et   al. \cite{Zhu2019UnsupervisedEgomotion} & No & Analog & 0.32 & 0.58 & 1.02 & 0.87 & \hl{-}  \\ \hline
        Zero   prediction & -  & - & 1.08 & 1.29 & 2.13 & 1.88  & \hl{-} \\ \hline
        \end{tabular}
        \end{table*}
        
        Optical flow is a computer vision task that aims to estimate the apparent movement of an object (pixel) in a sequence of images. Early works on this task were based on inputs from frame-based cameras. 
        However, these approaches suffered in high-speed motion and challenging lighting conditions while also incurring high energy and latency.
        We utilize inputs from a low-power asynchronous event-based camera that does not suffer from the above limitations.
        Recent works adopting an event-based approach to estimate optical flow, demonstrate their importance in terms of maintaining accuracy while being efficient at the same time \cite{Lee2020Spike-FlowNet:Networks, Zhu2018EV-FlowNet:Cameras,  Zhu2019UnsupervisedEgomotion}. 
        
        For example, \cite{Lee2020Spike-FlowNet:Networks} showed that combining SNN and ANN layers in a hybrid encoder-decoder architecture (Spike-FlowNet) could achieve a significant improvement in the computational efficiency of the model while achieving \hl{ } better performance than comparable ANN implementations (EV-FlowNet) \cite{Zhu2018EV-FlowNet:Cameras}. 
        Nevertheless, all previous works focused only on full precision implementations without exploring the potential losses due to hardware constraints.
        
        Here, we train our FSFN model, shown in Fig.~\ref{fig:spikeflownet}, on the Multi-Vehicle Stereo Event Camera (MVSEC) dataset \cite{Zhu2018ThePerception} following the hardware-aware training methodology discussed in Section~\ref{sec:hardware_aware}.
        For this purpose, we integrate our training methodology into the self-supervised pipeline training proposed in \cite{Lee2020Spike-FlowNet:Networks}.
        Similar to previous works, we train our models on the \textit{outdoor\_day2} sequence of the MVSEC dataset and evaluate on the \textit{outdoor\_day1} and \textit{indoor\_flying1,2,3} sequences.

        The quantitative results are shown in Table~\ref{table:mvsec_results}, where we report the Average End-point Error (AEE) metric compared with previous works.
        Here, the full precision FSFN (FSFN\textsubscript{FP}) gets better AEE performance than Spike-FlowNet and EV-FlowNet models, which have a similar number of parameters and training pipelines.
        The quantized FSFN model with HP-ADC (FSFN\textsubscript{HP-ADC}) shows performance comparable to FSFN\textsubscript{FP}. 
        This shows that our quantization-aware training can recover most of the performance of the full-precision models. 
        For ADC-Less training, we use the first mapping method discussed in Section~\ref{sec:adcless_aware} because optical flow requires less sparsity in the intermediate layers. 
        From Table~\ref{table:mvsec_results}, it can be seen that the performance of \hl{the} three ADC-Less models ($32$, $64$, and $128$) is superior to \cite{Zhu2018EV-FlowNet:Cameras}, and only slightly worse than \cite{Lee2020Spike-FlowNet:Networks}. 
        Note that the ADC-Less model trained with an xbar size of $64$ performs slightly better than the one using $32$. 
        The main reason \hl{that the} results \hl{ } do not follow an increasing trend is \hl{due to} the batch size used during the training. 
        The model with xbar size of $64$ allows using a larger batch size, which helps with the convergence of the training. 
        
\begin{figure}[!t]
        \centering
        \includegraphics[width=\columnwidth]{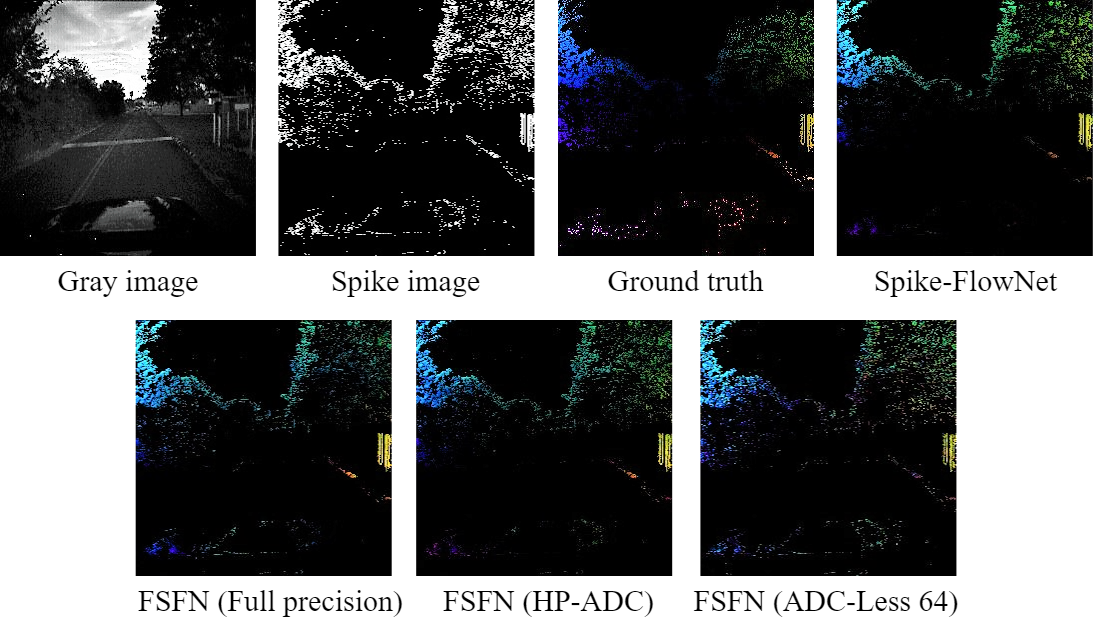}
        \caption{Masked optical flow evaluation and comparison with Spike-FlowNet. The sample is taken from \textit{outdoor\_day1} sequence, and the masked optical flow is the optical flow only in the pixels where there was an event, that is, using the spike image as the mask of the dense flow.}
        \label{fig:optical_flow}
        \end{figure}

        In addition, Fig.~\ref{fig:optical_flow} shows the masked optical flow produced by our models compared with the Spike-FlowNet and the ground truth sample in the \textit{outdoor\_day1} sequence. 
        As expected, the ADC-Less model produces a noisier estimation, but in general, it captures the apparent movement of different objects with a reasonable performance.
        The quantitative and qualitative results show that \hl{the} ADC-Less aware training scheme can be extended to complex spatio-temporal regression problems, like optical flow, and work well with self-supervised learning schemes.

    \subsection{Hardware Performance }\label{sec:hardware_performance}
    In this section, we estimate energy and latency improvements of the ADC-Less IMC architecture described in Section~\ref{sec:adcless_architecture} with respect to a conventional IMC architecture with HP-ADCs.
    For optical flow estimation (Section~\ref{sec:optical_flow}), we additionally compare the ADC-Less IMC with the results obtained for the FSFN running on an NVIDIA Jetson TX2 board. 
    
    \hl{We estimate the energy and latency of our architecture using} the DNN+NeuroSim~V1.3 \cite{Peng2019DNN+NeuroSim:Technologies} simulator.\hl{
    Specifically, we use NeuroSim to generate the chip floorplan for all the SNN architectures with different array sizes, i.e., estimate the total number of tiles and PE per tile.
    Also, for each chip floorplan, we used NeuroSim to estimate the energy and latency required to perform computations (using HP-ADC and ADC-Less crossbars) and those consumed by communication modules between layers during inference. 
    Note that the estimated energy and latency values for a regular ANN with binary activations are practically equivalent to those of an SNN for a single time-step.
    So we did not modify NeuroSim in any significant way as it can be expected that the estimation for an ANN with n-bit activations to be in the same order of magnitude as that of an SNN with ``n" time-steps.
    To complement those estimations, and since NeuroSim is not designed for spiking models, we designed and synthesized the digital LIF neuron (described in Section}~\ref{sec:digital_lif}\hl{) using ModelSim and Cadence RTL compiler using TSMC 65nm PDK. 
    The metrics obtained for an individual LIF neuron are $1448~\mu m^2$ of area, $1.202~mW$ of dynamic power, and $57.6~nW$ of leakage power. 
    Then following the tile structure shown in Fig.}~\ref{fig:imc_adcless}\hl{a and considering the per-layer speedup factor utilized in NeuroSim during simulation, we calculated the total number of LIF modules required per tile.  
    Using this information, we scaled the LIF module's energy and latency values for our analyses.

    The configuration used for simulation was as follows:} 1-bit ReRAM for bit-cells, Ron/Roff ratio of $150$, Flash ADCs, and $65~nm$ for the technology node.
    \hl{In addition, }for the ADC-Less architectures, we use a 1-bit ADC connected to each column similar to the sense amplifiers shown in Fig.~\ref{fig:imc_adcless}. This is feasible because of the small area required by 1-bit ADC. 
    In contrast, we use an 8-to-1 multiplexer to connect the HP-ADC, that is, $8$ columns bitlines share the same ADC\hl{. It would be unrealistic to use one HP-ADC per column due to large area requirement} \cite{Yu2020Compute-in-MemoryProspects, Peng2019DNN+NeuroSim:Technologies}. 
    For all the simulations, the resolution of the HP-ADC is set to 5-bit.
    \hl{Note that both HP-ADC IMC and ADC-Less IMC simulations use the same structure at Tile and PE level, with the main difference in the configuration of ADCs per crossbar. 
    
    In the following analysis, we focus on the potential improvement (ratio of magnitudes) of the ADC-Less architectures with respect to HP-ADC and not on the absolute magnitudes of energy and latency. }

        
        \subsubsection{Energy Improvements}\label{sec:energy_area}
        First, we analyze energy improvements for optical flow estimation.
        The improvements are computed as
        \begin{equation*}
            \mbox{improvement} = \frac{(\mbox{HP-ADC} - \mbox{ADC-Less})}{\mbox{ADC-Less}}
        \end{equation*}
        
\begin{figure}[!t]
        \centering
        \subfloat[\hl{Image classification (VGG16 and ResNet20)}]{\includegraphics[width=\columnwidth]{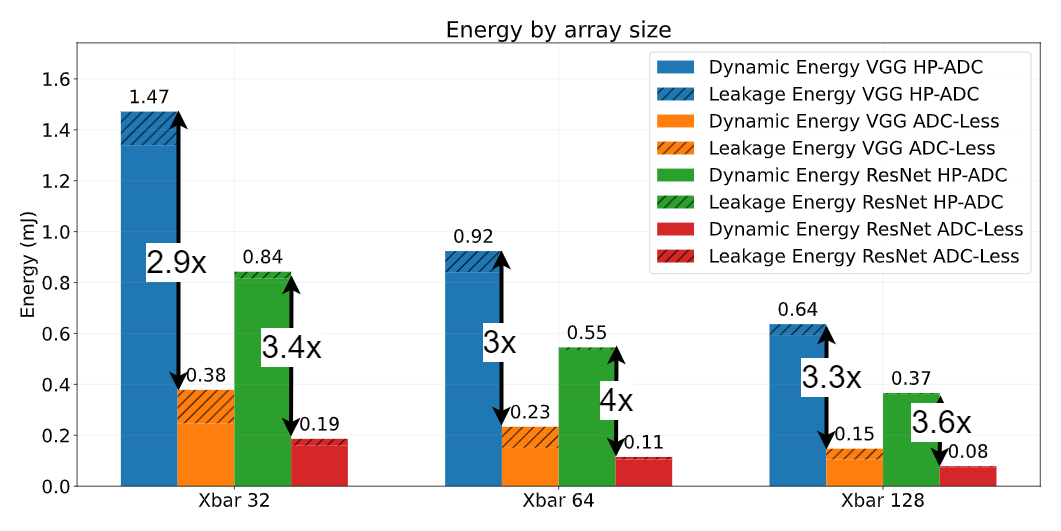}%
        \label{fig:energy_image}}
        
        \hfil
        \subfloat[Gesture recognition (DVSNet)]{\includegraphics[width=0.5\columnwidth]{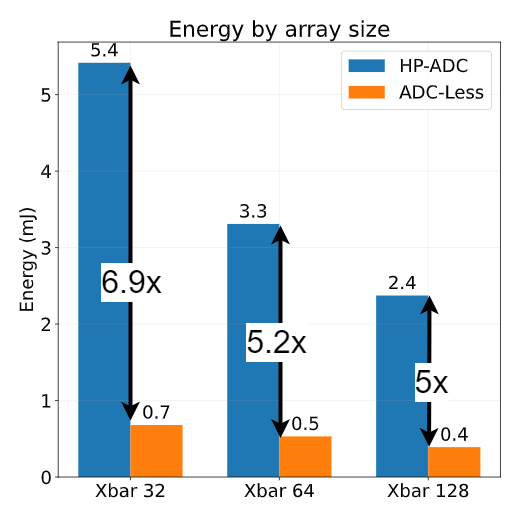}%
        \label{fig:energy_gesture}}
        \hfil
        \subfloat[\hl{Optical flow (FSFN)}]{\includegraphics[width=0.5\columnwidth]{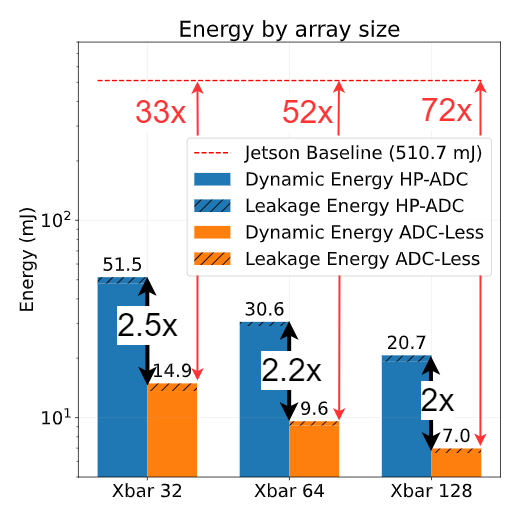}%
        \label{fig:energy_optical}}
        \caption{Energy consumption comparison of the ADC-Less and HP-ADC IMC architectures with different crossbar sizes, running different SNN models for different computer vision tasks.}
        \label{fig:energy_improvements}
        \end{figure}

        Fig.~\ref{fig:energy_optical} shows that the ADC-Less architecture consumes  \hl{$2-2.5\times$} lower energy than the same architecture with HP-ADC.
        As expected, when the crossbar array size increases, the energy consumed per inference \hl{ reduces}, and \hl{hence,} the gap between the ADC-Less and HP-ADC reduces\hl{.
        Note, however} the ADC-Less architecture still remains energy-efficient. 
        
        Since our main motivation is to enable optical flow estimation suitable for edge inference in real-time, we compare our results with the NVIDIA Jetson TX2 platform, an embedded system largely used to deploy DL models at the edge.
        We observe that the Jetson platform consumes $510.7~mJ$ per inference \hl{for} the FSFN model, making our ADC-Less architecture \hl{$33-72\times$} more energy-efficient, as depicted in Fig.~\ref{fig:energy_optical}.
        
        Similar results are obtained for gesture recognition ($5-6.9 \times $ lower energy consumption) and image classification (\hl{$2.9-4 \times $} lower energy consumption) tasks, as shown in Fig.~\ref{fig:energy_gesture} and Fig.~\ref{fig:energy_image} respectively.
        
        These results show that our proposed ADC-Less architecture is suitable as a low-energy inference engine for edge applications, allowing a direct trade-off between energy and accuracy \hl{with} crossbar array size.

        \subsubsection{Latency Improvements}\label{sec:latency}
        To estimate latency, we consider all the operations during a forward pass.
        The simulations were performed under a synchronous operation mode\hl{. 
        Thus} the clock period is determined by the compute-sense cycle. This is measured as the critical path from the input to the memory crossbar arrays till the ADC output generating the partial sum. The latency is determined by the total number of cycles needed for processing. 
        Given this configuration, the clock period for the ADC-Less architectures is around $1.1~ns$ while the HP-ADC clock period is around $11.9~ns$. 
        For all the cases, we report the latency under the assumption that it is possible to implement pipelined processing. 

\begin{figure}[!t]
        \centering
        \subfloat[Image classification (VGG16 and ResNet20)]{\includegraphics[width=\columnwidth]{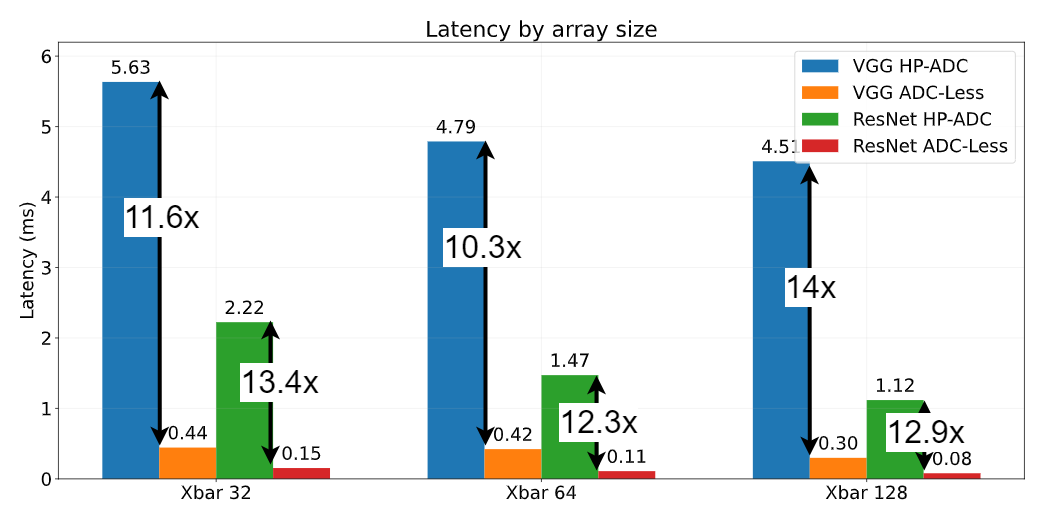}%
        \label{fig:latency_image}}
        \hfil
        \subfloat[Gesture recognition (DVSNet)]{\includegraphics[width=0.5\columnwidth]{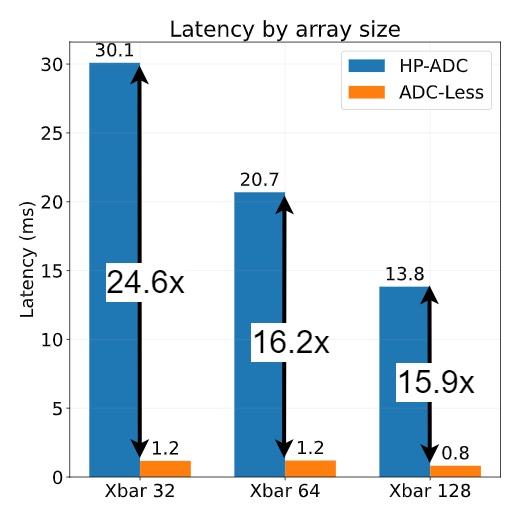}%
        \label{fig:latency_gesture}}
        \hfil
        \subfloat[Optical flow (FSFN)]{\includegraphics[width=0.5\columnwidth]{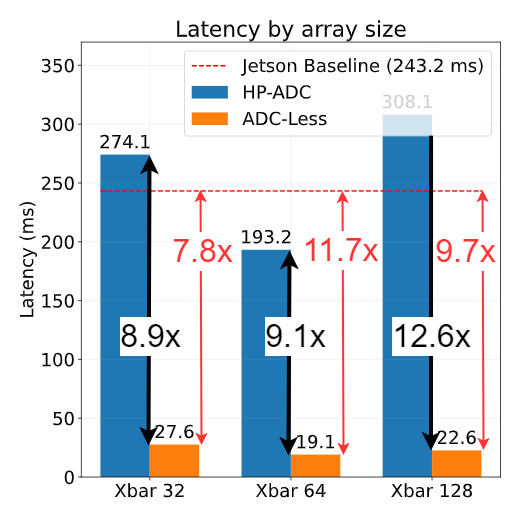}%
        \label{fig:latency_optical}}
        \caption{Latency comparison of the ADC-Less and HP-ADC IMC architectures with different crossbar sizes, running different SNN models for different computer vision tasks.}
        \label{fig:latency_improvements}
        \end{figure}

        For image classification, the ADC-Less \hl{architecture} presents latency improvement over the HP-ADC architecture \hl{($10.3-14\times$)}, as shown in Fig.~\ref{fig:latency_image}. 
        Similarly, gesture recognition experiments present latency improvements of $15.9-24.6\times$.
        In both cases, when the array size increase, the latency decrease.
        
        Similar to the previous section, for the optical flow estimation task, we compare the latency values of the ADC-Less architecture with the Jetson~TX2 platform and the HP-ADC.
        When the FSFN model is deployed on the Jetson platform, we obtain a high latency of $243.2~ms$, which is far from the \hl{ } real-time latency \hl{requirement} ($33~ms$ or $30$~FPS).  
        In contrast, as can be seen in Fig.~\ref{fig:latency_optical}, the ADC-Less architecture has a $7.8-11.7~\times$ latency improvement, which potentially enables real-time inference using FSFN.
        Similarly, the ADC-Less \hl{architecture} has an $8.9-12.6~\times$ latency improvement over the HP-ADC architecture.
        One important observation is that when the array size goes from $64$ to $128$, in Fig.~\ref{fig:latency_optical}, the latency increases\hl{.} 
        This \hl{is due to the fact that with increase in array size}, the model can be mapped using less number of crossbar arrays. This means that fewer tiles are required in the simulator \hl{($64$ tiles for the $64$ array size to $16$ tiles for the $128$ array size)}.
        \hl{The} reduction in the number of tiles produces a bottleneck that increases the latency\hl{.
        This} did not happen for other cases, as shown in Fig.~\ref{fig:latency_improvements}\hl{,} because the \hl{other} models are smaller than FSFN.
        

\section{Conclusion}


Spiking Neural Networks are suitable for performing complex sequential tasks \hl{if we can effectively use} their membrane potential as short-term memory. 
Note, however, most commercial hardware accelerators (GPU-based) are not optimized for such networks.
While crossbar-based IMC architectures can be energy-efficient, the usage of expensive HP-ADC is a major obstacle.
Unilaterally reducing the ADC precision (and its associated cost) results in significant degradation in the model's performance. To recover the performance we developed hardware\hl{-}aware training that enables IMC architectures with ADC precision as low as 1-bit.
The proposed approach shows significant energy and latency improvements, $2-7\times$ and $8.9-24.6\times$ respectively, with respect to HP-ADC IMC architectures with comparable accuracy.
Moreover, in contrast to previous works, our methodology naturally extends the range of applications beyond simple image classification tasks to more challenging sequential tasks such as optical flow estimation and gesture recognition.
\ifCLASSOPTIONcompsoc
  \section*{Acknowledgments}
\else
  \section*{Acknowledgment}
\fi

This work was supported in part by the Micro4AI program from IARPA, the Center for Brain-Inspired Computing (C-BRIC),
one of six centers in JUMP, funded by Semiconductor Research Corporation (SRC) and DARPA, the National Science Foundation, and Intel Corporation.

\ifCLASSOPTIONcaptionsoff
  \newpage
\fi



\bibliographystyle{IEEEtran}
\bibliography{IEEEabrv, ./references.bib}
%



%

\begin{IEEEbiography}[{\includegraphics[width=1in,height=1.25in,clip,keepaspectratio]{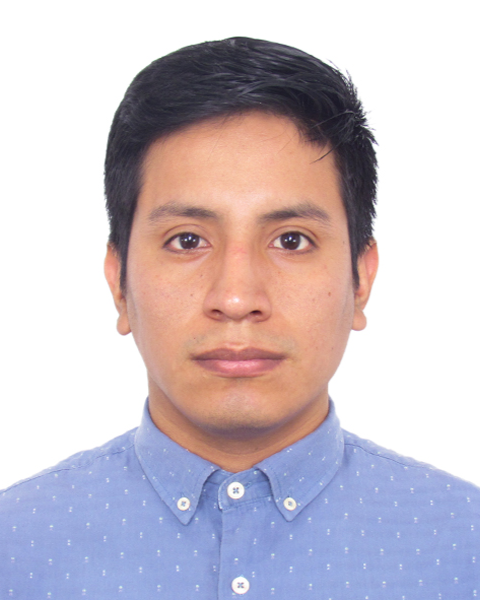}}]{Marco~P.~E.~Apolinario}
received his B.S. degree in Electronics Engineering from the National University of Engineering (UNI), Lima, Peru, in 2017. Currently, he is pursuing his Ph.D. degree at Purdue University under the guidance of Prof. Kaushik Roy. His research interests include hardware-software co-design for brain-inspired computing, in-memory computing architectures, and learning algorithms for spiking neural networks.
\end{IEEEbiography}

\begin{IEEEbiography}[{\includegraphics[width=1in,height=1.25in,clip,keepaspectratio]{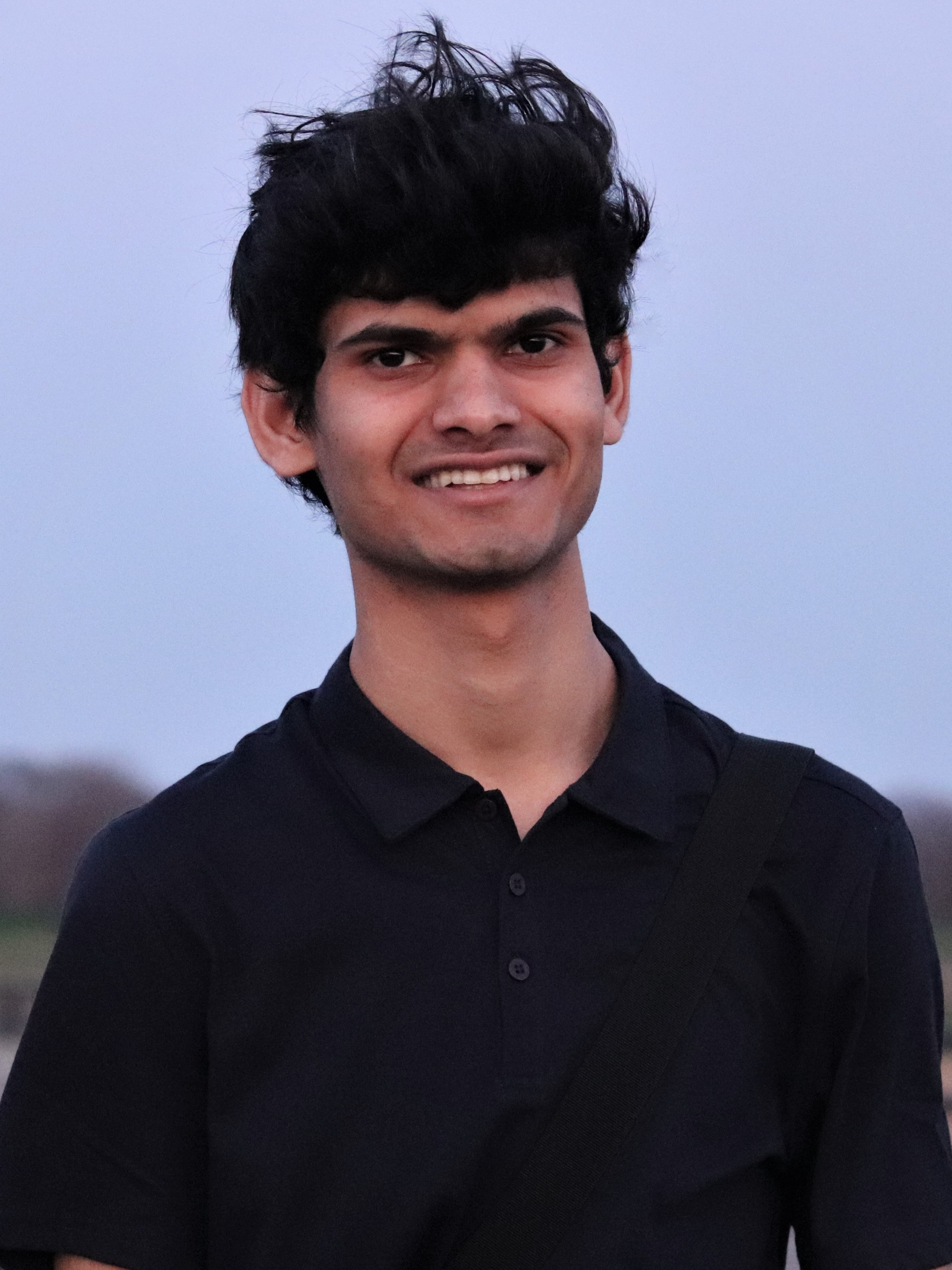}}]{Adarsh~Kumar~Kosta}
received his Dual degree (Integrated B.Tech + M.Tech) in Electronics and Electrical Communication Engineering from Indian Institute of Technology Kharagpur, India, in 2018. He is currently pursuing his Ph.D. degree at Purdue University under the guidance of Prof. Kaushik Roy. His research interests lie in neuromorphic computing, in-memory computing architectures and hardware-aware algorithms for deep learning.
\end{IEEEbiography}

\begin{IEEEbiography}[{\includegraphics[width=1in,height=1.25in,clip,keepaspectratio]{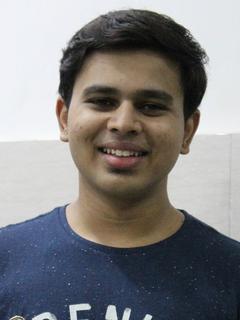}}]{Utkarsh~Saxena}
Utkarsh received his B.Tech degree in Electrical Engineering (Power and Automation) from Indian Institute of Technology (IIT), Delhi in 2019. Currently, he is pursuing his PhD degree under the guidance of Prof. Kaushik Roy. His research interests include designing algorithms and architectures for deep learning systems using CMOS and various post-CMOS devices.
\end{IEEEbiography}

\begin{IEEEbiography}[{\includegraphics[width=1in,height=1.25in,clip,keepaspectratio]{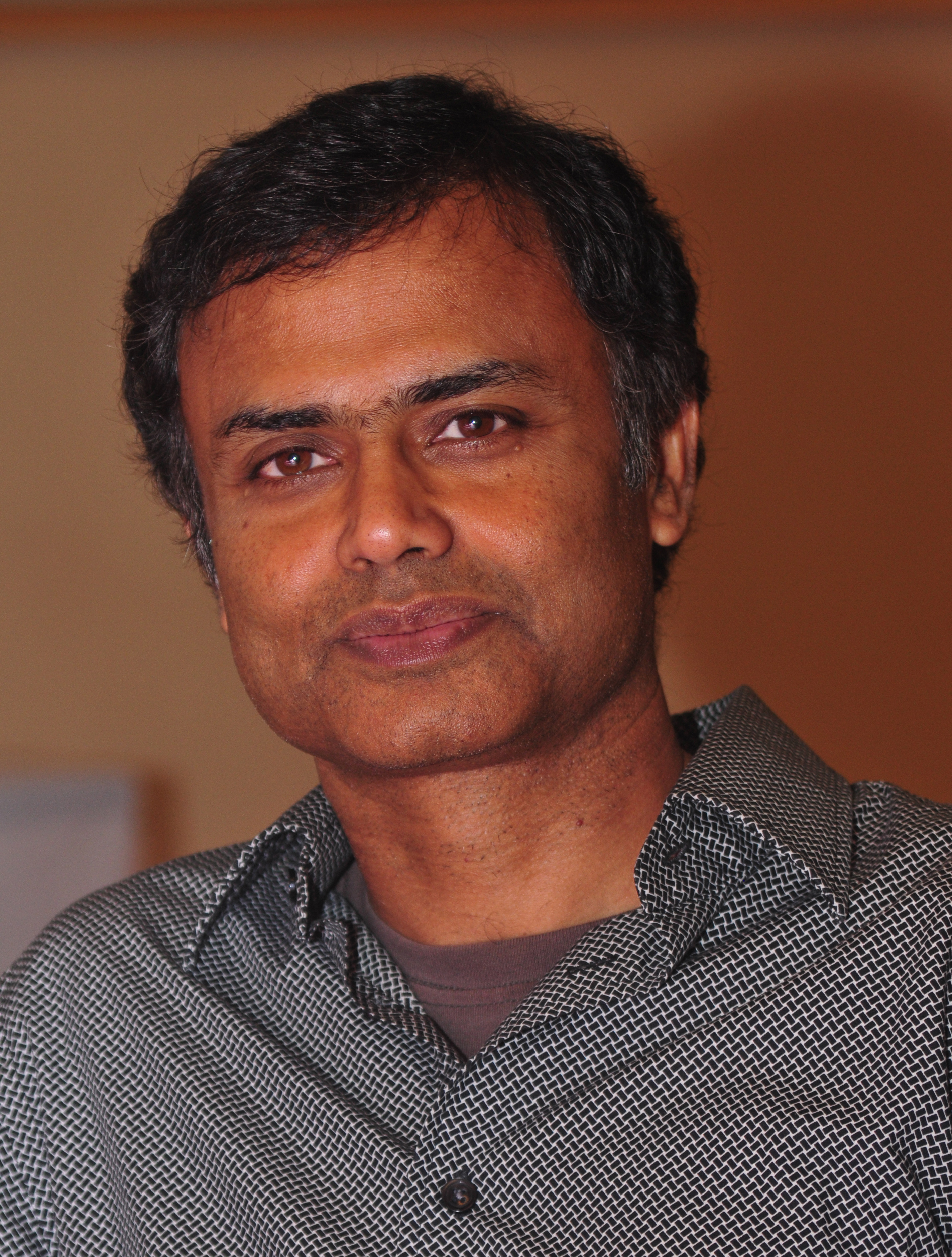}}]{Kaushik~Roy}
is the Edward G. Tiedemann, Jr., Distinguished Professor of Electrical and Computer Engineering at Purdue University. He received his BTech from Indian Institute of Technology, Kharagpur, PhD from University of Illinois at Urbana-Champaign in 1990 and joined the Semiconductor Process and Design Center of Texas Instruments, Dallas, where he worked for three years on FPGA architecture development and low-power circuit design. His current research focuses on cognitive algorithms, circuits and architecture for energy-efficient neuromorphic computing/ machine learning, and neuro-mimetic devices. Kaushik has supervised 99 PhD dissertations and his students are well placed in universities and industry. He is the co-author of two books on Low Power CMOS VLSI Design (John Wiley \& McGraw Hill).
\end{IEEEbiography}





\end{document}